\documentclass{article}

\PassOptionsToPackage{dvipsnames}{xcolor}

\usepackage{iclr}

\iclrfinalcopy

\usepackage{algorithm}
\usepackage{algpseudocode}
\usepackage{amsmath}
\usepackage{amssymb}
\usepackage{booktabs}
\usepackage{colortbl}
\usepackage{enumitem}
\usepackage{graphicx}
\usepackage{makecell}
\usepackage{marvosym}
\usepackage{multirow}
\usepackage{pifont}
\usepackage{setspace}
\usepackage{tabularx}
\usepackage{tcolorbox}
\usepackage{times}
\usepackage{xurl}
\usepackage{wrapfig}

\definecolor{citecolor}{HTML}{1F7ACE}
\usepackage[colorlinks,citecolor=citecolor]{hyperref}

\renewcommand{\paragraph}[1]{\textbf{#1}\hspace{2mm}}

\newcommand{\cmark}{\ding{51}}
\newcommand{\xmark}{\ding{55}}

\newcommand{\ie}{\textit{i.e.}}
\newcommand{\eg}{\textit{e.g.}}

\definecolor{mcolor}{HTML}{4088F3}
\definecolor{icolor}{HTML}{FFA21B}
\definecolor{ncolor}{HTML}{FF7E6F}
\definecolor{dcolor}{HTML}{22CA60}

\definecolor{plannercolor}{HTML}{E2EDFE}
\definecolor{groundercolor}{HTML}{FEF8DD}
\definecolor{verifiercolor}{HTML}{FCEBEC}
\definecolor{answerercolor}{HTML}{E1F2E7}
\definecolor{promptcolor}{HTML}{F3F3F5}

\title{\vspace{-4.15mm}\raisebox{-1.65mm}{\includegraphics[scale=0.045]{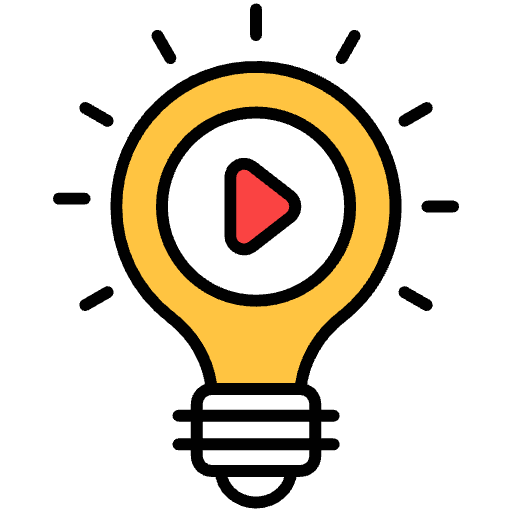}}Video{\color{mcolor} M}{\color{icolor} i}{\color{ncolor} n}{\color{dcolor} d}: A Chain-of-LoRA Agent for \\ Temporal-Grounded Video Reasoning}

\author{Ye~Liu$^1$\thanks{Equal contribution.\quad\textsuperscript{\Letter}Corresponding authors.}\enspace, Kevin~Qinghong~Lin$^{2*}$, Chang~Wen~Chen\textsuperscript{$1$\,\Letter}, Mike~Zheng~Shou\textsuperscript{$2$\,\Letter} \\
$^1$The Hong Kong Polytechnic University $^2$National University of Singapore \\
\texttt{coco.ye.liu@connect.polyu.hk}}

\begin{document}

\maketitle

\begin{abstract}
Videos, with their unique temporal dimension, demand precise grounded understanding, where answers are directly linked to visual, interpretable evidence. Despite significant breakthroughs in text-based reasoning with large language models, multi-modal reasoning -- especially for videos -- remains limited. In this work, we fill this gap by introducing \textbf{VideoMind}, a novel video-language agent for temporal-grounded video reasoning. Our method involves two key innovations: (1) We identify four essential capabilities for grounded video reasoning and propose a role-based agentic workflow, comprising a \texttt{planner} to coordinate roles, a \texttt{grounder} for temporal event localization, a \texttt{verifier} to assess event candidates, and an \texttt{answerer} for question answering. (2) To efficiently integrate these roles during inference, we propose a novel \textbf{Chain-of-LoRA} mechanism, where a unified base model with multiple LoRA adapters is leveraged to enable seamless role switching, balancing efficiency and flexibility. Extensive experiments on 15 benchmarks across Grounded VideoQA, Video Temporal Grounding, and General VideoQA tasks demonstrate the effectiveness of the proposed scheme in advancing video agent, test-time scaling, and long-form video reasoning. Code, models, datasets, and demos are available at \url{https://videomind.github.io/}.
\end{abstract}

\section{Introduction}

Recent advancements in large language models (LLMs) have demonstrated remarkable success in text-based reasoning \citep{cot,tot,reflexion}, significantly improving both accuracy and interpretability in complex problem-solving scenarios \citep{react}. Following these breakthroughs, efforts have been devoted to extending these reasoning capabilities to multi-modal domains \citep{seed2.0,llavacot,llamavo1} such as vision-centric science \citep{scienceqa} and math \citep{visionmath} understanding.

\begin{figure}
\centering
\includegraphics[width=0.995\linewidth]{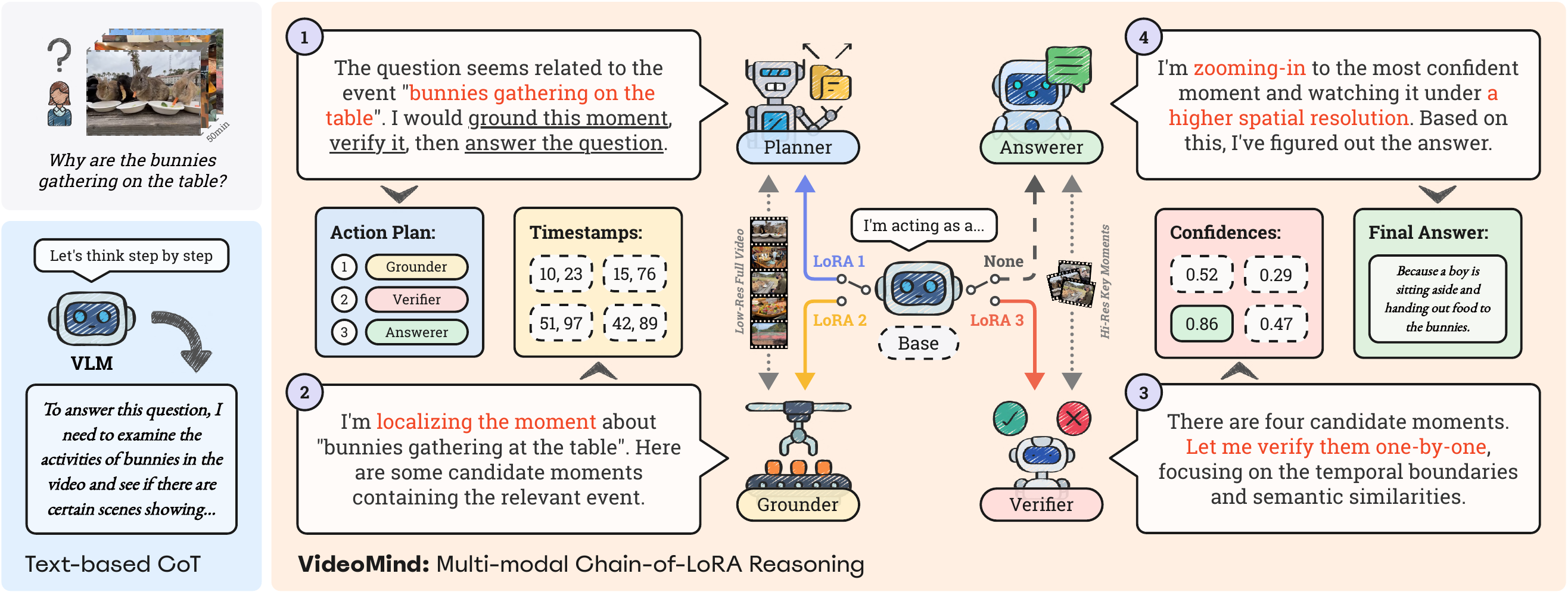}
\vspace{-5mm}
\caption{Illustration of VideoMind's Chain-of-LoRA reasoning mechanism. The problem is decomposed by the planner and distributed to the grounder, verifier, and answerer to systematically localize, verify, and interpret the relevant video moments.}
\label{fig:teaser}
\vspace{-3mm}
\end{figure}

Among multi-modal signals, videos pose a unique challenge due to their temporal dimension, introducing complexities absent in images or text. Effective video reasoning requires not only recognizing visual appearances but also understanding how they evolve over time \citep{nextgqa,cgbench,etbench,vtgsurvey}. While recent visual Chain-of-Thought (CoT) methods \citep{mmcot,llavacot,llamavo1} excel at generating detailed thoughts for static images, they struggle with long videos as they cannot explicitly localize or revisit earlier parts of the sequence, as presented in Figure~\ref{fig:teaser}~(left). Humans, by contrast, can reason over long videos with ease \citep{singlepulsenc,singlepulse,visualthalamus}: they break down complex problems, identify relevant moments, revisit them to confirm details, and synthesize their observations into coherent answers. This natural proficiency motivates the development of an AI agent that emulates this process -- flexibly coordinating multiple capabilities to achieve advanced, vision-centric reasoning.

In this work, we introduce \textbf{VideoMind}, a video-language agent with enhanced temporal-grounded reasoning capabilities. To meet the demands of diverse tasks, we define four essential roles for understanding complex long-form videos: (1) a {\color{mcolor}\texttt{\textbf{planner}}} to decompose tasks and coordinate other roles, (2) a {\color{icolor}\texttt{\textbf{grounder}}} for precise moment localization, (3) a {\color{ncolor}\texttt{\textbf{verifier}}} for moment candidates assessment, and (4) an {\color{dcolor}\texttt{\textbf{answerer}}} for moment-aware response generation. Each role is carefully designed to deliver strong performance, for example, the grounder is equipped with a timestamp decoder to ensure accurate temporal grounding. To enable efficient integration of these roles, we also propose a novel \textbf{Chain-of-LoRA} mechanism, where all the roles are implemented based on a unified LMM backbone with role-specific LoRA adapters \citep{lora}. Therefore, role-specific capabilities can be trained separately on tailored datasets. During inference, all the LoRA parameters are cached into the memory, so that each role could be activated by simply switching to the corresponding LoRA, as shown in Figure~\ref{fig:teaser}~(right). This approach reflects a minimalist yet flexible design philosophy, facilitating seamless transitions and interactions among roles without incurring the memory overhead of maintaining multiple full models. As a result, VideoMind achieves both efficiency and flexibility on diverse video understanding tasks.

We conduct extensive experiments on 15 public benchmarks, including 4 on Grounded VideoQA, 6 on Video Temporal Grounding, and 5 on General VideoQA, to evaluate the effectiveness of our approach. VideoMind exhibits strong adaptability in addressing diverse reasoning tasks by jointly providing accurate responses and temporal-grounded evidence. Notably, our 2B model surpasses GPT-4o \citep{gpt4o} and Gemini-1.5-Pro \citep{gemini1.5} on several long video benchmarks such as CG-Bench \citep{cgbench}, MLVU \citep{mlvu}, and LVBench \citep{lvbench}. State-of-the-art performance is also achieved on temporal grounding datasets including QVHighlights \citep{qvhighlights} and Charades-STA \citep{charadessta}. We further conduct ablation studies to justify our design choices, particularly the Chain-of-LoRA mechanism for enhancing flexibility while preserving efficiency. Our contributions are summarized as follows:
\begin{enumerate}
\item We propose \textbf{VideoMind}, a multi-modal agentic framework that enhances video reasoning by emulating human cognitive processes, including task decomposition, moment localization and verification, and answer synthesis. It addresses the unique challenges of long video reasoning in a progressive and structured manner.
\item We introduce \textbf{Chain-of-LoRA}, an efficient test-time scaling mechanism that enables a single model to seamlessly switch among multiple roles. This approach enhances VideoMind's flexibility without incurring additional memory overhead.
\item Our method demonstrates strong performance across three scenarios: Grounded VideoQA, Video Temporal Grounding, and General VideoQA. Notably, VideoMind-2B outperforms GPT-4o and Gemini-1.5-Pro on several long video benchmarks.
\end{enumerate}

\section{Related Work}

\paragraph{Temporal-grounded Video Understanding} Significant advances in video understanding have propelled tasks such as video retrieval \citep{egovlp,vlog} and captioning \citep{lavila,movieseq,videollmol}. However, these models often lack \textit{visually grounded correspondence} and interpretability, particularly for long-form videos. The task of Video Temporal Grounding \citep{charadessta,activitynetcaptions} tackles this issue by requiring precise temporal localization for diverse queries, though regression-based models \citep{umt,r2tuning} excel at localization but fall short in providing textual interpretability. Recent benchmarks \citep{nextgqa,cgbench,etbench} intensify this challenge, demanding both reasoning for complex questions and fine-grained temporal correspondence. Previous baselines for these tasks typically rely on multi-task objectives or modular agents composed of distinct components \citep{videoagentlong,videoagentmem}, often yielding sub-optimal performance or overly complex systems, which constrain their efficiency and flexibility. Our VideoMind is an agentic workflow built upon a unified LMM, seamlessly integrating multiple functionalities while enhancing localization and interpretability, thus surpassing the limitations of prior methods.

\paragraph{Multi-modal Reasoning} Large Multi-modal Models \citep{llava,unipixel} exhibit generalized capabilities such as free-form question answering. However, they fall short in addressing complex challenges that often require reasoning \citep{cot}. One approach to overcome this is to develop agent-based interfaces \citep{llovi,langrepo}, which integrates textual outputs from visual tools to enable reasoning via LLMs. Advanced methods \citep{vipergpt,mmreact,assistgpt} invoke visual APIs through progressive execution and reasoning. Alternatively, pure text-based reasoning \citep{openaio1,deepseekr1} has been a dominant paradigm in LLMs, exemplified by training with long CoT processes using reinforcement learning, which provides detailed step-by-step reasoning, with some works \citep{longvilar1,videor1} extending this mechanism to the visual domain. Despite these advances, extending reasoning to videos remains an open challenge. Given the long-context nature of informative videos, we believe that \textit{a vision-centric} CoT should incorporate a human-like \citep{shortsinglepulse,eegfmri,intraandinter,rsfmrieeg} re-watching strategy and self-validation of intermediate observations, leading us to introduce a novel Chain-of-LoRA framework for video reasoning.

\paragraph{Inference-time Searching} Inference-time searching has emerged as a critical technique for tackling complex reasoning challenges in different domains. The advent of OpenAI o1 \citep{openaio1} has advanced these inference-time techniques within LLMs by integrating sampling strategies such as controlled decoding \citep{transferqstar}, Best-of-N sampling \citep{letsverify}, and Monte Carlo Tree Search (MCTS) \citep{llmmcts}, allowing LLMs to iteratively refine outputs and achieve superior performance without altering their underlying weights. However, the potential of inference-time searching remains largely untapped in video understanding, where temporal reasoning introduces unique challenges. In our framework, we explore how such a strategy can be tailored for video temporal reasoning, observing that models are highly sensitive to the selection of temporal segments, often producing unreliable predictions when segment choices are sub-optimal. To address this, we propose a \textit{moment-level} searching approach where a grounder generates multiple candidates, followed by a verifier that evaluates and determines the correct correspondence. The framework also supports flexible role switching with minimal memory overhead.

\section{Method}\label{sec:method}

\begin{figure}
\centering
\includegraphics[width=0.95\linewidth]{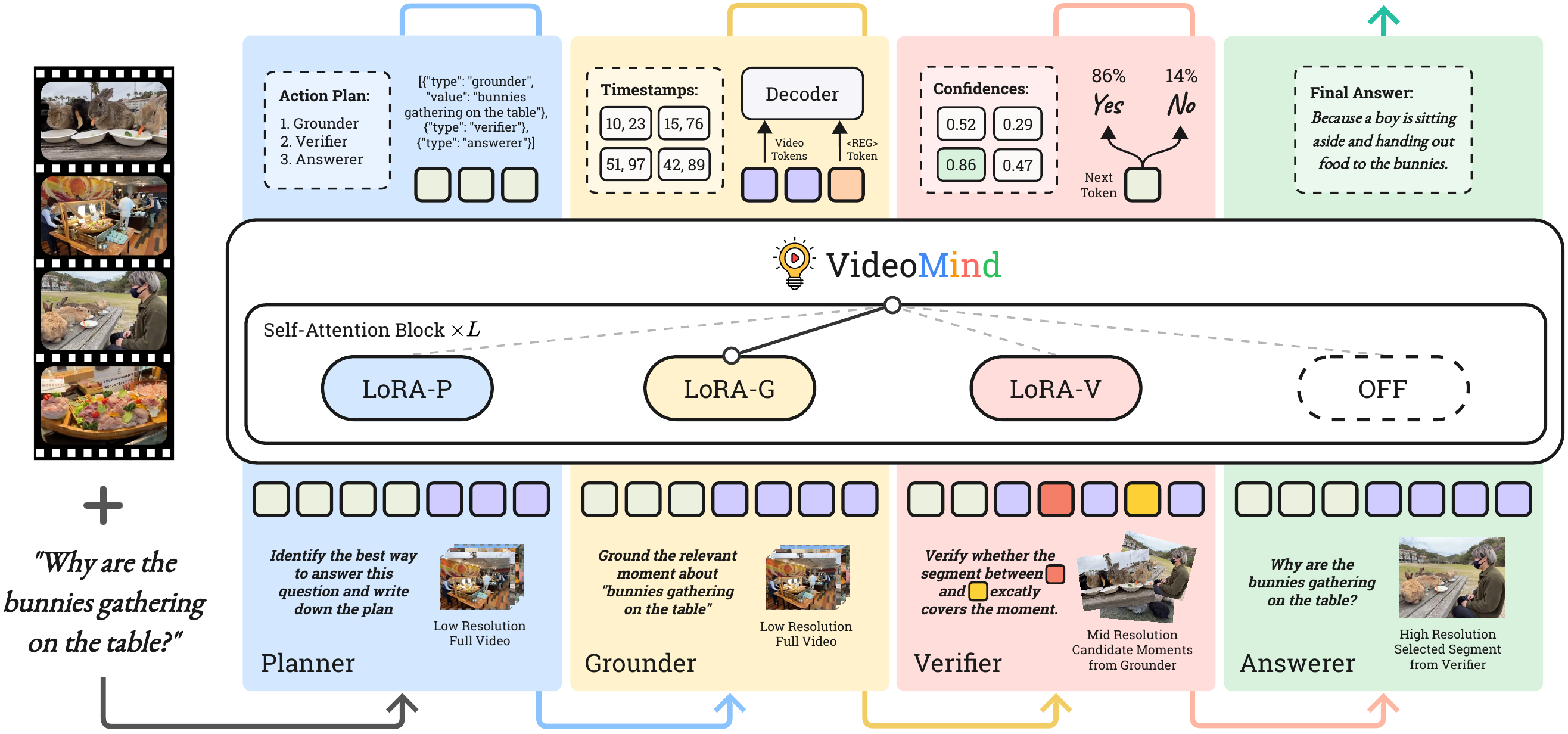}
\vspace{-2mm}
\caption{The overall workflow of VideoMind. Given a video and a query, it adaptively activates different roles (\eg, \texttt{Planner} $\to$ \texttt{Grounder} $\to$ \texttt{Verifier} $\to$ \texttt{Answerer} in this case) and performs step-by-step reasoning by calling individual modules.}
\label{fig:method}
\vspace{-1mm}
\end{figure}

\paragraph{Overview} Figure~\ref{fig:method} provides an overview of VideoMind. Our model derives from the Qwen2-VL \citep{qwen2vl} architecture, consisting of an LLM backbone and a ViT-based visual encoder support dynamic resolution inputs. Given a video input $\mathcal{V}$ and a text query $\mathcal{Q}$, the model performs step-by-step reasoning by adaptively calling different roles: (1) \texttt{\textbf{{\color{mcolor} Planner}}}: Dynamically coordinates the following roles based on the query. (2) \texttt{\textbf{{\color{icolor} Grounder}}}: Identifies and localizes relevant video moments. (3) \texttt{\textbf{{\color{ncolor} Verifier}}}: Evaluates the validity of the moments identified by the grounder, refining them through a zoom-in process with boolean outputs. (4) \texttt{\textbf{{\color{dcolor} Answerer}}}: Generates the final response in natural language. This mechanism enables the models to \textbf{revisit the videos several times} (with varying temporal segments \& spatial resolutions) to derive the final response.

\vspace{-1mm}
\subsection{Planner}
\vspace{-1mm}

\begin{figure}
\centering
\includegraphics[width=0.65\linewidth]{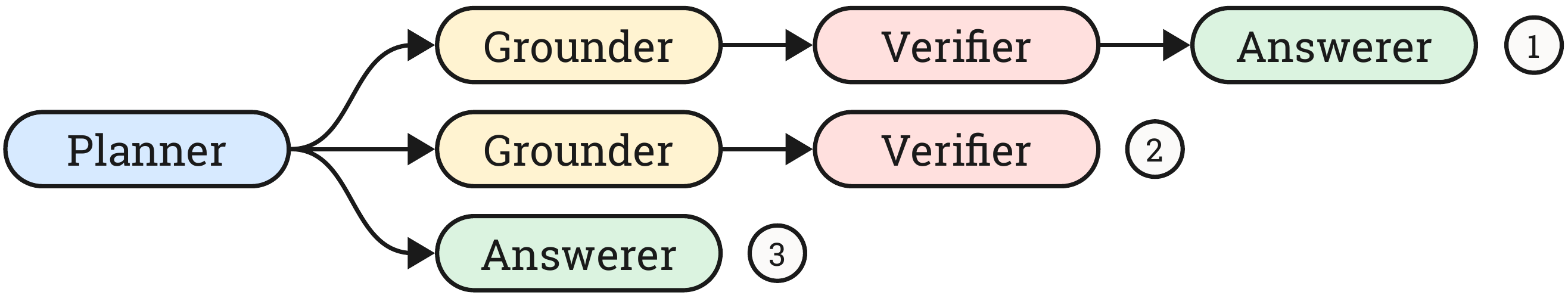}
\vspace{-2mm}
\caption{Planner coordinates all the other roles based on the video and query context, offering three reasoning plans and a query rephrasing mechanism to address diverse demands.}
\label{fig:planner}
\vspace{-3mm}
\end{figure}

An agent must be flexible enough to handle diverse tasks and efficiently determine which functions (roles) to call. To achieve this, we design the \texttt{\textbf{planner}}, which dynamically coordinates all the other roles for each query. It decides the sequence of function calls based on the multi-modal context. We utilize a JSON-style object \texttt{\{{\color{NavyBlue} "type"}:\,{\color{Orange} "<role>"},\,{\color{NavyBlue} "value"}:\,{\color{Orange} "<argument>"}\}} to denote a function call. In this way, a sequence of roles can be succinctly represented as a list of such objects. Three reasoning plans for different tasks are pre-defined and illustrated in Figure~\ref{fig:planner}.

\textbf{(1) Grounding \& Verifying \& Answering:} This plan requires the agent to generate both a textual response and a corresponding temporal moment. For example, in Grounded VideoQA scenarios \citep{nextqa}, to answer the question \textit{``What is the boy doing when the baby is crying?''}, the agent should identify the moment of \textit{``baby is crying''}, and then investigate the boy's activity.

\textbf{(2) Grounding \& Verifying:} This plan is designed for grounding-only tasks such as moment retrieval \citep{qvhighlights,charadessta}. For questions like \textit{``When does the woman go downstairs?''}), the model should provide precise timestamps directly as the answer. Since the grounding results could potentially be unreliable, an extra zoom-in verification step is necessary.

\textbf{(3) Answering Only:} If the question is straightforward (\eg, \textit{``Summarize this video''}) or the video is very short (\eg, less than 10s), it could be unnecessary to perform grounding. Instead, the model should watch the entire video and answer the question directly.

\paragraph{Query Rephrasing} When the user query lacks sufficient detail for accurate moment localization, the planner is allowed to \textbf{rephrase} the question into a more descriptive version. For instance, the question \textit{``What is the person sitting on the bed doing as the baby plays?''} may confuse the grounder as it contains multiple events (\textit{``person sitting on the bed''} and \textit{``baby plays''}). It can be rephrased to \textit{``the baby is playing''} as an accurate scene description.

To train the planning and query rephrasing capabilities, we curated a dataset of 39K samples (shown in Table~\ref{tab:training}) from public benchmarks. For planning, we aligned each reasoning plan with corresponding question types: \textit{temporal} questions from NExT-QA \citep{nextqa} are assigned to Plan-1, moment queries from QVHighlights \citep{qvhighlights} are for Plan-2, and \textit{causal} \& \textit{descriptive} questions from NExT-QA \citep{nextqa} are for Plan-3. For query rephrasing, we leverage GPT-4o mini \citep{gpt4o} to generate synthetic \texttt{video} + \texttt{question} $\to$ \texttt{query} samples for training.

\vspace{-1mm}
\subsection{Grounder}
\vspace{-1mm}

The \texttt{\textbf{grounder}} aims to localize relevant moments (\ie, predicting start and end timestamps) based on text queries, thereby supporting the reasoning process by identifying visual cues. This requirement calls for the development of an LMM with robust temporal grounding capabilities.

\paragraph{Timestamp Decoder} Instead of directly predicting timestamps through language modeling \citep{timechat} or special tokens \citep{vtimellm,etbench}, we develop a timestamp decoder to maximize the LMM-based grounding performance. Specifically, we introduce a \texttt{<REG>} token to facilitate this process. When the \texttt{<REG>} token is generated, the last-layer hidden states of it and all the visual tokens will be sent into the decoder for timestamp prediction, obtaining a tuple $[t_{start}, t_{end}]$ representing the normalized start and end timestamps.

\begin{wrapfigure}{r}{0.375\textwidth}
\vspace{-4.25mm}
\centering
\includegraphics[width=0.95\linewidth]{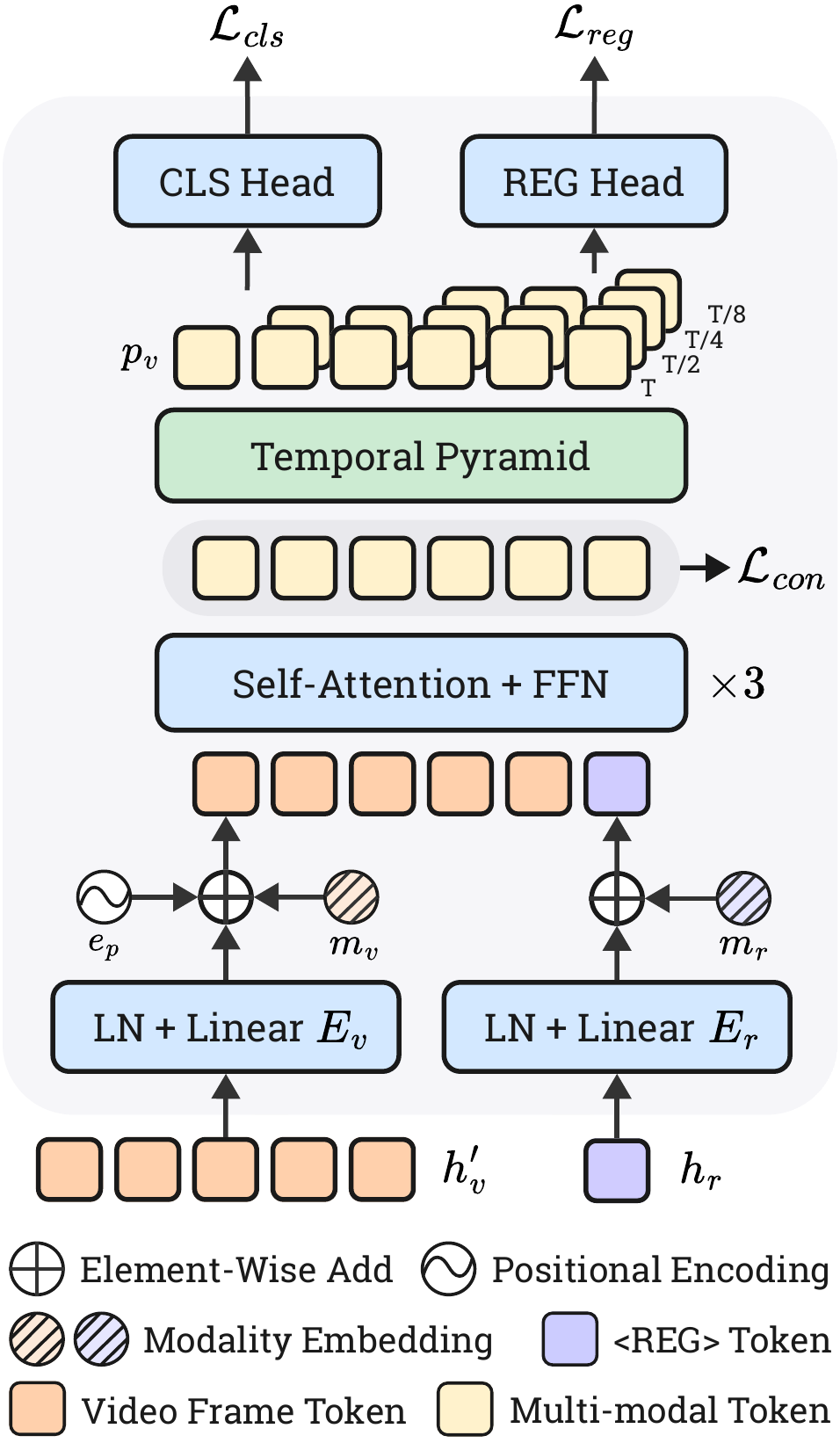}
\vspace{-2mm}
\caption{Detailed architecture of the timestamp decoder.}
\label{fig:decoder}
\vspace{5mm}
\end{wrapfigure}

As shown in Figure~\ref{fig:decoder}, the decoder accepts the hidden states of the visual tokens $\mathbf{h}_{v} \in \mathbb{R}^{(T \times H \times W) \times D_{L}}$ and the \texttt{<REG>} token $\mathbf{h}_{r} \in \mathbb{R}^{1 \times D_{L}}$ as inputs, where $T$, $H$, $W$, $D_{L}$ are the downsampled number of frames, height, width, and hidden dimensions of the LLM, respectively. We apply a 1D average pooling with kernel size and stride equal to $H \times W$ to compress the visual tokens to one token per frame.
\begin{gather}
\mathbf{h}_{v}' = \mathrm{AvgPool}(\mathbf{h}_{v}) \in \mathbb{R}^{T \times D_{L}}
\end{gather}
Then, $\mathbf{h}_{v}'$ and $\mathbf{h}_{r}$ are projected by two linear layers $E_{v}$ and $E_{r}$ to reduce the hidden dimension to $D$.
\begin{gather}
\mathbf{e}_{v} = E_{v}(\mathbf{h}_{v}') \in \mathbb{R}^{T \times D},\hspace{2mm}\mathbf{e}_{r} = E_{r}(\mathbf{h}_{r}) \in \mathbb{R}^{1 \times D}
\end{gather}
The resulting $\mathbf{e}_{v}$ and $\mathbf{e}_{r}$ serve as consolidated representations of the video frames and the query\footnote{We use the term ``query'' to denote the features of \texttt{<REG>} token.}, respectively. To effectively integrate their information, we concatenate them along the sequence dimension and send them into a three-layer transformer encoder \citep{transformer}.
\begin{gather}
[\mathbf{e}_{v}'; \mathbf{e}_{r}'] = \mathrm{Transformer}([\mathbf{e}_{v} + \mathbf{m}_{v} + \mathbf{e}_{p}; \mathbf{h}_{r} + \mathbf{m}_{r}])
\end{gather}
Here, modality indicators $m_v \in \mathbb{R}^{1 \times D}$ and $m_r \in \mathbb{R}^{1 \times D}$ are randomly initialized learnable embeddings. $m_v$ is expanded to $T \times D$ before being added with $e_v$. $e_{p}$ is a normalized sinusoidal positional encoding \citep{transformer} for preserving temporal awareness. The output sequence is split back into $\mathbf{e}_{v}'$ and $\mathbf{e}_{r}'$, indicating the contextualized frame and query embeddings, respectively.

\paragraph{Temporal Feature Pyramid} To improve the model's adaptability to videos and moments of varying lengths, we map $\mathbf{e}_{v}'$ into a four-level temporal feature pyramid \citep{r2tuning,actionformer}. Each level is produced by a \texttt{Conv1D\,$\to$\,LayerNorm\,$\to$\,SiLU} block, where the \texttt{Conv1D} employs a kernel size and stride of 2. Therefore, the resulting four levels retain $1$, $1/2$, $1/4$, and $1/8$ of the original sequence length, respectively. To accelerate the prediction, we concatenate the sequences from all pyramid levels along the temporal dimension to form $\mathbf{p}_{v}$ with length $L = T + T/2 + T/4 + T/8$, allowing parallelized prediction across temporal resolutions.

\paragraph{Prediction Heads} We introduce two heads for timestamps prediction: \textbf{(1) A classification head} is designed for frame-level foreground-background classification. This is instantiated by a two-layer \texttt{Conv1D} module with kernel size 3 and padding 1, followed by a $\mathrm{Sigmoid}$ activation. The outputs are frame-level confidence scores $\{\hat{c}_i\}_{i=0}^{L}$ indicating whether each frame falls inside the desired moment. A binary focal loss \citep{retinenet} is utilized to optimize these scores.
\begin{gather}
\mathcal{L}_{cls} = -\lambda_{cls}\alpha(1 - \hat{c}_i)^\gamma\log(\hat{c}_i)
\end{gather}
Here, $\alpha = 0.9$ and $\gamma = 2.0$ are hyperparameters of the focal loss, and $\lambda_{cls}$ is the loss reweighing term. \textbf{(2) A boundary regression head} is adopted to predict the frame-level temporal offsets for start and end boundaries $\{[\hat{b}_i^s, \hat{b}_i^e]\}_{i=0}^{L}$. This is also a two-layer \texttt{Conv1D} block (with 2 output channels), followed by an exponential activation. Predictions from different pyramid levels are further modulated by different learnable scaling factors. These outputs are supervised by an $L1$ loss.
\begin{gather}
\mathcal{L}_{reg} = \lambda_{reg}(|b_i^s - \hat{b}_i^s| + |b_i^e - \hat{b}_i^e|)
\end{gather}
In order to realize better alignment between $e'_v$ and $e'_r$, we incorporate an additional contrastive loss to encourage learning more discriminative representations. Specifically, we calculate the cosine similarities among all frame-query pairs (denoted as $\{s_i\}_{i=0}^{L}$), then sample a positive frame (falling within the ground truth boundary) and apply the following optimization objective:
\begin{gather}
\mathcal{L}_{con} = -\lambda_{con}\log\frac{\exp(s_p/\tau)}{\exp(s_p/\tau) + \sum_{i \in \Theta}\exp(s_i/\tau)}
\end{gather}
Here, $\Theta$ is the set of frame indices with $s_p > s_i$, and $\tau = 0.07$ is the temperature parameter. The final loss for the timestamp decoder is the sum of these losses at all layers with $\lambda_{cls} = 5.0$, $\lambda_{reg} = 1.0$, and $\lambda_{con} = 0.05$. The training datasets for the grounder are listed in Table~\ref{tab:training}.

\begin{table}
\centering
\scriptsize
\setlength{\tabcolsep}{5pt}
\caption{Training datasets for different roles. Source datasets were repurposed for training planner and verifier. \textit{mr} and \textit{step} denote the moment retrieval and step localization subsets, respectively.}
\vspace{2mm}
\begin{tabularx}{0.915\linewidth}{l|c|l}
\toprule
\textbf{Role} & \textbf{\#Samples} & \textbf{Source Datasets} \\
\midrule
\texttt{\textbf{Planner}} & 39K & NeXT-QA (34K), QVHighlights (5K) \\
\midrule
\multirow{1.9}{*}{\texttt{\textbf{Grounder}}} & \multirow{1.9}{*}{210K} & QVHighlights (5K), DiDeMo (33K), TACoS (9K), InternVid-VTime (54K), CosMo-Cap (87K), \\
&& QuerYD (19K), HiREST$_{\textit{mr}}$ (8K), HiREST$_{\textit{step}}$ (4K) \\
\midrule
\texttt{\textbf{Verifier}} & 232K & DiDeMo (165K), TACoS (43K), QVHighlights (24K) \\
\bottomrule
\end{tabularx}
\label{tab:training}
\vspace{-3mm}
\end{table}

\vspace{-1mm}
\subsection{Verifier}
\vspace{-1mm}

Key moments are crucial for providing visual cues, yet they might be unreliable due to grounding errors. Thus, further verifications are necessary. We let the grounder generate top-5 predictions, then employ the \texttt{\textbf{verifier}} to select the most reliable one. This process is presented below.

\begin{wrapfigure}{r}{0.425\textwidth}
\vspace{-2mm}
\centering
\includegraphics[width=\linewidth]{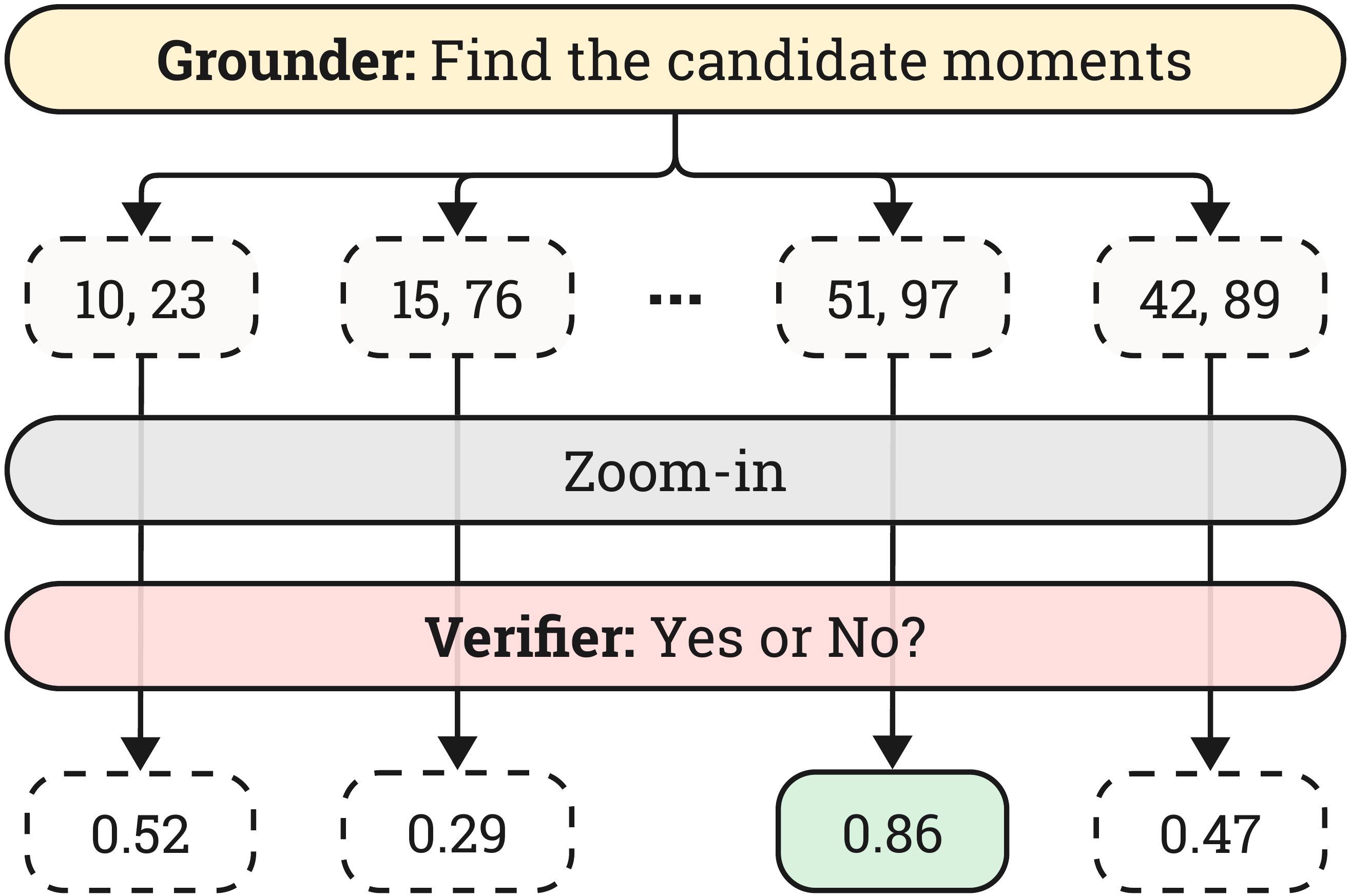}
\vspace{-5.5mm}
\caption{The grounder generates multiple candidate moments, which are then refined by the verifier via \textbf{zooming-in} to investigate and select the best one.}
\label{fig:verifier}
\vspace{-5mm}
\end{wrapfigure}

\paragraph{Recap by Zooming-in} For each candidate moment, we apply a zoom-in strategy by expanding the boundaries by 50\% on both sides and temporally cropping the enlarged segment. The resulting segment and the original text query are sent to the verifier to assess whether the queried event exactly occurs within the temporal boundaries. To enhance boundary awareness, we adopt two special tokens, \texttt{<SEG-START>} and \texttt{<SEG-END>}, to explicitly mark the beginning and end of the moment. These tokens are inserted among the visual tokens at the corresponding frames, effectively guiding the model in recognizing moment boundaries.

\paragraph{Boolean Judgement} The verifier's responses are binary, \ie, either \textit{``Yes''} or \textit{``No''}. To train this role, we sample predictions from the grounder and assign binary labels based on an IoU threshold of 0.5. The model is then fine-tuned via SFT to predict these labels. During inference, for each candidate moment, we employ teacher forcing to obtain the likelihoods of the \texttt{<Yes>} and \texttt{<No>} tokens, denoted as $L_y$ and $L_n$, respectively. The confidence score is then computed as $\mathrm{Sigmoid}(L_y - L_n)$. The moment with the highest score is selected and passed to the answerer.

\vspace{-1mm}
\subsection{Answerer}
\vspace{-1mm}

The \texttt{\textbf{answerer}} responds to the given question based on the cropped video segment (w/ grounder) or the whole video (w/o grounder). Since the objective of this role is strictly aligned with existing LMMs, we employ the original model directly \textit{without fine-tuning or architectural modifications.}

\begin{table}
\centering
\scriptsize
\setlength{\tabcolsep}{5pt}
\caption{Performance comparison on Grounded VideoQA on CG-Bench \citep{cgbench}.}
\vspace{2mm}
\begin{tabularx}{0.7235\linewidth}{l|c|c|cccc}
\toprule
\textbf{Method} & \textbf{Size} & \textbf{long-acc.} & \textbf{mIoU} & \textbf{rec.@IoU} & \textbf{acc.@IoU} \\
\midrule
GPT-4o \citep{gpt4o} & -- & \textbf{45.2} & 5.62 & 8.30 & \underline{4.38} \\
Gemini-1.5-Pro \citep{gemini1.5} & -- & 37.2 & 3.95 & 5.81 & 2.53 \\
Claude-3.5-Sonnet \citep{claude3.5sonnet} & -- & 40.5 & 3.99 & 5.67 & 2.79 \\
\midrule
Video-LLaVA \citep{videollava} & 7B & 16.2 & 1.13 & 1.96 & 0.59 \\
VideoLLaMA \citep{videollama} & 7B & 18.4 & 1.21 & 1.87 & 0.84 \\
VideoChat2 \citep{mvbench} & 7B & 19.3 & 1.28 & 1.98 & 0.94 \\
ST-LLM \citep{stllm} & 7B & 23.8 & 2.23 & 2.86 & 1.13 \\
ShareGPT4Video \citep{sharegpt4video} & 16B & 26.7 & 1.85 & 2.65 & 1.01 \\
Chat-UniVi-v1.5 \citep{chatunivi} & 13B & 25.9 & 2.07 & 2.53 & 1.21 \\
VILA \citep{vila} & 8B & 28.7 & 1.56 & 2.89 & 1.35 \\
LongVA \citep{longva} & 7B & 28.7 & 2.94 & 3.86 & 1.78 \\
LLaVA-OneVision \citep{llavaonevision} & 7B & 31.1 & 1.63 & 1.78 & 1.08 \\
Video-CCAM \citep{videoccam} & 14B & 29.7 & 2.63 & 3.48 & 1.83 \\
Kangaroo \citep{kangaroo} & 8B & 30.2 & 2.56 & 2.81 & 1.94 \\
VITA \citep{vita} & 8$\times$7B & 33.3 & 3.06 & 3.53 & 2.06 \\
Qwen2-VL \citep{qwen2vl} & 72B & 41.3 & 3.58 & 5.32 & 3.31 \\
InternVL2 \citep{internvl2} & 78B & \underline{42.2} & 3.91 & 5.05 & 2.64 \\
\midrule
\rowcolor{blue!7.5} \textbf{VideoMind} (Ours) & 2B & 31.0 & \underline{5.94} & \underline{8.50} & 4.02 \\
\rowcolor{blue!7.5} \textbf{VideoMind} (Ours) & 7B & 38.4 & \textbf{7.10} & \textbf{9.93} & \textbf{4.67} \\
\bottomrule
\end{tabularx}
\label{tab:cgbench}
\vspace{-4mm}
\end{table}

\begin{table}
\centering
\scriptsize
\setlength{\tabcolsep}{5pt}
\caption{Performance comparison on Grounded VideoQA on ReXTime \citep{rextime}. \underline{FT} indicates fine-tuning on the target dataset.}
\vspace{2mm}
\begin{tabularx}{0.755\linewidth}{l|cc|ccccc}
\toprule
\textbf{Method} & \textbf{Size} & \textbf{FT} & \textbf{R@0.3} & \textbf{R@0.5} & \textbf{mIoU} & \textbf{Acc} & \textbf{Acc@IoU} \\
\midrule
VTimeLLM \citep{vtimellm} & 7B & \xmark & 28.84 & 17.41 & 20.14 & 36.16 & -- \\
TimeChat \citep{timechat} & 7B & \xmark & 14.42 & 7.61 & 11.65 & 40.04 & -- \\
LITA \citep{lita} & 13B & \xmark & 29.49 & 16.29 & 21.49 & 34.44 & -- \\
\midrule
\textcolor{gray}{VTimeLLM \citep{vtimellm}} & \textcolor{gray}{7B} & \textcolor{gray}{\cmark} & \textcolor{gray}{43.69} & \textcolor{gray}{26.13} & \textcolor{gray}{29.92} & \textcolor{gray}{57.58} & \textcolor{gray}{17.13} \\
\textcolor{gray}{TimeChat \citep{timechat}} & \textcolor{gray}{7B} & \textcolor{gray}{\cmark} & \textcolor{gray}{40.13} & \textcolor{gray}{21.42} & \textcolor{gray}{26.29} & \textcolor{gray}{49.46} & \textcolor{gray}{10.92} \\
\midrule
\rowcolor{blue!7.5} \textbf{VideoMind} (Ours) & 2B & \xmark & \underline{34.31} & \underline{22.69} & \underline{24.83} & \underline{69.06} & \underline{17.26} \\
\rowcolor{blue!7.5} \textbf{VideoMind} (Ours) & 7B & \xmark & \textbf{38.22} & \textbf{25.52} & \textbf{27.61} & \textbf{74.59} & \textbf{20.20} \\
\bottomrule
\end{tabularx}
\label{tab:rextime}
\vspace{-4mm}
\end{table}

\vspace{-1mm}
\subsection{Chain-of-LoRA}
\vspace{-1mm}

The four roles introduced above demonstrate distinct yet complementary capabilities, collaborating to achieve advanced vision-centric reasoning. However, simply integrating these roles into a single model poses challenges, as their core functionalities can interfere with one another. To avoid inefficiently implementing them as multiple models while still accommodating diverse demands, we propose a novel \textbf{Chain-of-LoRA} mechanism to enable flexible and efficient role switching.

In greater detail, all roles are based on a shared LMM backbone and are augmented with different LoRA adapters \citep{lora}. Note that an additional timestamp decoder is used exclusively by the grounder. During inference, the framework dynamically activates role-specific LoRA adapters according to the planner, thereby maximizing the strengths of each role while minimizing the memory consumption and architectural modifications to the base model.

\vspace{-1mm}
\section{Experiments}
\vspace{-1mm}

We evaluate the effectiveness of VideoMind through extensive experiments across 15 public benchmarks. Specifically, we study the following research questions.
\begin{enumerate}[label=\textbf{Q\arabic*.},topsep=0pt,itemsep=0pt,parsep=2pt]
\item Whether VideoMind is flexible and effective on diverse video understanding tasks compared to the corresponding baselines with task-specific designs?
\item Compared with (1) training a single agent on multiple tasks or (2) distributing all roles to different models, what advantages does Chain-of-LoRA offer?
\item What effects does each individual design contribute? More importantly, whether each role is necessary for building such a video reasoning system?
\end{enumerate}

Detailed information about the benchmarks, evaluation settings, implementation details, and more experimental results can be found in the appendix.

\begin{table}
\centering
\scriptsize
\setlength{\tabcolsep}{5pt}
\caption{Performance comparison on Grounded VideoQA on NExT-GQA \citep{nextgqa}.}
\vspace{2mm}
\begin{tabularx}{0.9075\linewidth}{l|c|ccc|ccc|c}
\toprule
\multirow{2.6}{*}{\textbf{Method}} & \multirow{2.6}{*}{\textbf{Size}} & \multicolumn{3}{c|}{\textbf{IoU}} & \multicolumn{3}{c|}{\textbf{IoP}} & \multirow{2.6}{*}{\textbf{Acc@GQA}} \\
\cmidrule{3-5} \cmidrule{6-8}
&& R@0.3 & R@0.5 & mIoU & R@0.3 & R@0.5 & mIoP \\
\midrule
FrozenBiLM NG+ \citep{frozenbilm} & 890M & 13.5 & 6.1 & 9.6 & 28.5 & 23.7 & 24.2 & 17.5 \\
SeViLA \citep{sevila} & 4B & 29.2 & 13.8 & 21.7 & 34.7 & 22.9 & 29.5 & 16.6 \\
LangRepo \citep{langrepo} & 8$\times$7B & -- & 12.2 & 18.5 & -- & 28.7 & 31.3 & 17.1 \\
VideoStreaming \citep{videostreaming} & 8.3B & -- & 13.3 & 19.3 & -- & 31.0 & 32.2 & 17.8 \\
LLoVi \citep{llovi} & 1.8T & -- & 15.3 & 20.0 & -- & \textbf{36.9} & \underline{37.3} & 24.3 \\
HawkEye \citep{hawkeye} & 7B & 37.0 & 19.5 & 25.7 & -- & -- & -- & -- \\
VideoChat-TPO \citep{videochattpo} & 7B & 41.2 & \underline{23.4} & 27.7 & 47.5 & 32.8 & 35.6 & \underline{25.5} \\
\midrule
\rowcolor{blue!7.5} \textbf{VideoMind} (Ours) & 2B & \underline{45.2} & 23.2 & \underline{28.6} & \underline{51.3} & 32.6 & 36.4 & 25.2 \\
\rowcolor{blue!7.5} \textbf{VideoMind} (Ours) & 7B & \textbf{50.2} & \textbf{25.8} & \textbf{31.4} & \textbf{56.0} & \underline{35.3} & \textbf{39.0} & \textbf{28.2} \\
\bottomrule
\end{tabularx}
\label{tab:nextgqa}
\vspace{-4mm}
\end{table}

\begin{table}
\centering
\scriptsize
\setlength{\tabcolsep}{3.85pt}
\caption{Performance comparison on video temporal grounding on Charades-STA \citep{charadessta} and ActivityNet-Captions \citep{activitynetcaptions}. \underline{FT} means fine-tuning on the target dataset.}
\vspace{2mm}
\begin{tabularx}{0.955\linewidth}{l|cc|cccc|cccc}
\toprule
\multirow{2.6}{*}{\textbf{Method}} & \multirow{2.6}{*}{\textbf{Size}} & \multirow{2.6}{*}{\textbf{FT}} & \multicolumn{4}{c|}{\textbf{Charades-STA}} & \multicolumn{4}{c}{\textbf{ActivityNet-Captions}} \\
\cmidrule{4-11}
&&& R@0.3 & R@0.5 & R@0.7 & mIoU & R@0.3 & R@0.5 & R@0.7 & mIoU \\
\midrule
VTimeLLM \citep{vtimellm} & 7B & \xmark & 51.0 & 27.5 & 11.4 & 31.2 & \underline{44.0} & \underline{27.8} & \underline{14.3} & \underline{30.4} \\
TimeChat \citep{timechat} & 7B & \xmark & 51.5 & 32.2 & 13.4 & -- & -- & -- & -- & -- \\
Momentor \citep{momentor} & 7B & \xmark & 42.6 & 26.6 & 11.6 & 28.5 & 42.9 & 23.0 & 12.4 & 29.3 \\
ChatVTG \citep{chatvtg} & 7B & \xmark & 52.7 & 33.0 & 15.9 & 34.9 & 40.7 & 22.5 & 9.4 & 27.2 \\
VideoChat-TPO \citep{videochattpo} & 7B & \xmark & 58.3 & 40.2 & 18.4 & 38.1 & -- & -- & -- & -- \\
E.T. Chat \citep{etbench} & 4B & \xmark & 65.7 & 45.9 & 20.0 & 42.3 & 24.1 & 12.8 & 6.1 & 18.9 \\
Grounded-VideoLLM \citep{groundedvideollm} & 4B & \xmark & 54.2 & 36.4 & 19.7 & 36.8 & -- & -- & -- & -- \\
TRACE \citep{trace} & 7B & \xmark & -- & 40.3 & 19.4 & -- & -- & -- & -- & -- \\
LLaVA-ST \citep{llavast} & 7B & \xmark & 63.1 & 44.8 & 23.4 & 42.4 & -- & -- & -- & -- \\
UniTime \citep{unitime} & 7B & \xmark & -- & \textbf{59.1} & \textbf{31.9} & \textbf{52.2} & -- & 22.8 & 14.1 & 27.3 \\
\midrule
\rowcolor{blue!7.5} \textbf{VideoMind} (Ours) & 2B & \xmark & \underline{67.6} & \underline{51.1} & 26.0 & 45.2 & \underline{44.0} & 26.5 & 12.6 & 30.1 \\
\rowcolor{blue!7.5} \textbf{VideoMind} (Ours) & 7B & \xmark & \textbf{73.5} & \textbf{59.1} & \underline{31.2} & \underline{50.2} & \textbf{48.4} & \textbf{30.3} & \textbf{15.7} & \textbf{33.3} \\
\bottomrule
\end{tabularx}
\label{tab:vtg}
\vspace{-4mm}
\end{table}

\vspace{-1mm}
\subsection{Q1: Comparison with State-of-the-Arts}
\vspace{-1mm}

\paragraph{Grounded Video Question Answering} Table~\ref{tab:cgbench} compares the Grounded VideoQA performance on CG-Bench \citep{cgbench}, a challenging video benchmark with an average duration of 27 minutes. On temporal grounding metrics (mIoU and rec.@IoU), our lightweight 2B model outperforms all the baselines, including GPT-4o \citep{gpt4o} and Gemini 1.5 Pro \citep{gemini1.5}. Our 7B model further setups a new state-of-the-art on clue-grounded QA (acc.@IoU). In Table~\ref{tab:rextime} and Table~\ref{tab:nextgqa}, we further present the comparison results on ReXTime \citep{rextime} and NExT-GQA
\begin{wraptable}{l}{0.605\textwidth}
\vspace{-3.5mm}
\scriptsize
\setlength{\tabcolsep}{4.4pt}
\caption{Performance comparison on General VideoQA on Video-MME \citep{videomme}, MLVU \citep{mlvu}, and LVBench \citep{lvbench}.}
\vspace{2mm}
\begin{tabularx}{\linewidth}{l|c|cc|c|c}
\toprule
\multirow{2.6}{*}{\textbf{Method}} & \multirow{2.6}{*}{\textbf{Size}} & \multicolumn{2}{c|}{\textbf{Video-MME}} & \textbf{MLVU} & \textbf{LVBench} \\
\cmidrule{3-4} \cmidrule{5-5} \cmidrule{6-6}
&& All & Long & M-Avg & Overall \\
\midrule
\textcolor{gray}{GPT-4o \citep{gpt4o}} & \textcolor{gray}{--} & \textcolor{gray}{71.9} & \textcolor{gray}{65.3} & \textcolor{gray}{54.5} & \textcolor{gray}{30.8} \\
\textcolor{gray}{Gemini-1.5-Pro \citep{gemini1.5}} & \textcolor{gray}{--} & \textcolor{gray}{75.0} & \textcolor{gray}{67.4} & \textcolor{gray}{--} & \textcolor{gray}{33.1} \\
\midrule
Video-LLaVA \citep{videollava} & 7B & 41.1 & 37.8 & 29.3 & -- \\
TimeChat \citep{timechat} & 7B & 34.3 & 32.1 & 30.9 & 22.3 \\
MovieChat \citep{moviechat} & 7B & 38.2 & 33.4 & 25.8 & 22.5 \\
PLLaVA \citep{pllava} & 34B & 40.0 & 34.7 & 53.6 & 26.1 \\
VideoChat-TPO \citep{videochattpo} & 7B & 48.8 & 41.0 & 54.7 & -- \\
LongVA \citep{longva} & 7B & 52.6 & 46.2 & 56.3 & -- \\
\midrule
\rowcolor{blue!7.5} \textbf{VideoMind} (Ours) & 2B & \underline{55.4} & \underline{46.3} & \underline{58.7} & \underline{35.4} \\
\rowcolor{blue!7.5} \textbf{VideoMind} (Ours) & 7B & \textbf{58.2} & \textbf{49.2} & \textbf{64.4} & \textbf{40.8} \\
\bottomrule
\end{tabularx}
\label{tab:general}
\vspace{-3mm}
\end{wraptable}
\citep{nextgqa}. Despite the challenges posed by the causal event relationships on ReXTime, our model can successfully identify the correct moment, resulting in significant performance boosts compared with zero-shot baselines. On NExT-GQA, compared to agent-based solutions such as LLoVi \citep{llovi} and LangRepo \citep{langrepo} and end-to-end methods like VideoChat-TPO \citep{videochattpo}, VideoMind demonstrates its effectiveness on both key event grounding and question answering.

\paragraph{Video Temporal Grounding} We also evaluate the grounder and verifier on video temporal grounding datasets. The results on Charades-STA \citep{charadessta} and ActivityNet-Captions \citep{activitynetcaptions} are shown in Table~\ref{tab:vtg}. Benefiting from (1) the timestamp decoder design, and (2) a verifier that refines the results by focusing on critical segments, our model surpasses all LLM-based temporal grounding methods and yields competitive results compared to fine-tuned experts.

\paragraph{General Video Question Answering} We are also interested in whether our temporally augmented design can improve general VideoQA tasks. In Table~\ref{tab:general}, we evaluate our model on three long video benchmarks to determine if the Chain-of-LoRA design generalizes to common settings. Our designs effectively help the model localize cue segments before answering the question.

\begin{figure}[t]
\centering
\includegraphics[width=1.0\linewidth]{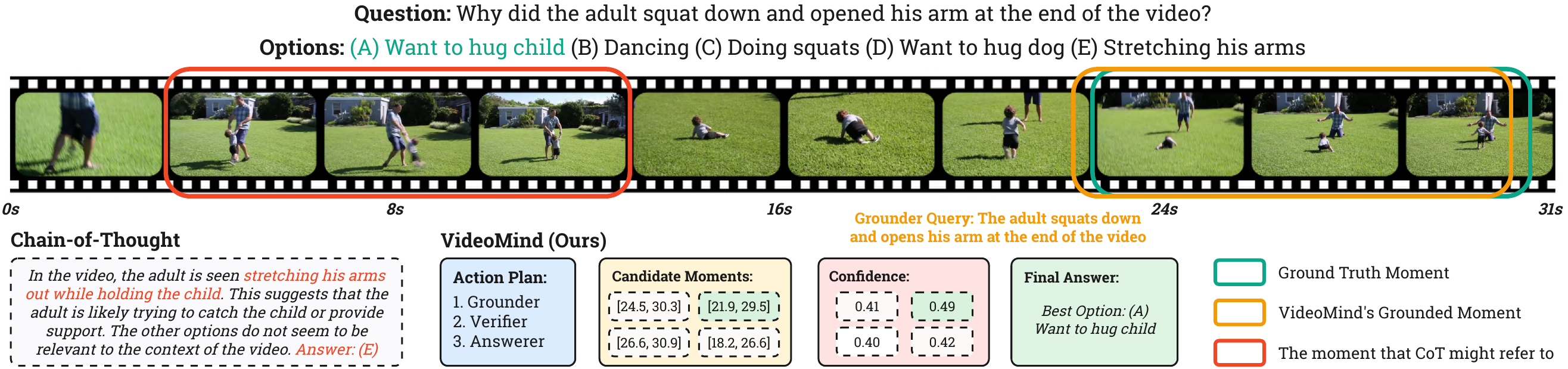}
\vspace{-5.5mm}
\caption{Visualization of the reasoning process of VideoMind. Through chaining the \texttt{planner}, \texttt{grounder}, \texttt{verifier}, and \texttt{answerer}, our model accurately localizes the critical moment and selects the correct answer, avoiding confusion from incorrect segments.}
\label{fig:vis0}
\vspace{-4mm}
\end{figure}

\begin{table}
\centering
\scriptsize
\begin{minipage}{0.5315\textwidth}
\setlength{\tabcolsep}{2.35pt}
\caption{\setstretch{1.115} Performance and efficiency comparison of different test-time scaling and role integration strategies. \underline{Mem} indicates the peak GPU memory consumption. Notably, Chain-of-LoRA achieves the best performance with minimal memory cost.}
\vspace{2mm}
\begin{tabularx}{\linewidth}{lc|cc|cc|cc}
\toprule
\multirow{2.6}{*}{\textbf{Method}} & \multirow{2.6}{*}{\textbf{Mem}} & \multicolumn{2}{c|}{\textbf{NExT-GQA}} & \multicolumn{2}{c|}{\textbf{Charades-STA}} & \multicolumn{2}{c}{\textbf{Video-MME}} \\
\cmidrule{3-4} \cmidrule{5-6} \cmidrule{7-8}
&& mIoU & Acc & R@0.5 & mIoU & All & Long \\
\midrule
Qwen2-VL-2B & 4.1G & -- & 69.6 & -- & -- & 53.0 & 43.1 \\
+\,CoT & 4.1G & -- & 69.7 & -- & -- & 52.8 & 43.3 \\
\midrule
+\,All-in-One & 4.2G & \underline{28.0} & \underline{70.5} & \underline{47.8} & \underline{42.1} & \underline{53.6} & \underline{43.6} \\
+\,All-Distributed & 16.6G & \textcolor{gray}{28.6} & \textcolor{gray}{71.4} & \textcolor{gray}{51.1} & \textcolor{gray}{45.2} & \textcolor{gray}{55.4} & \textcolor{gray}{46.3} \\
\rowcolor{blue!7.5} +\,\textbf{Chain-of-LoRA} & \textbf{4.2G} & \textbf{28.6} & \textbf{71.4} & \textbf{51.1} & \textbf{45.2} & \textbf{55.4} & \textbf{46.3} \\
\bottomrule
\end{tabularx}
\label{tab:chain}
\end{minipage}
\hfill
\begin{minipage}{0.4485\textwidth}
\setlength{\tabcolsep}{2.75pt}
\caption{Effects of individual roles. \underline{A}, \underline{G}, \underline{V}, \underline{P}, \underline{G\%} denote the answerer, grounder, verifier, planner, and the percentage of samples processed with the grounder, respectively.}
\vspace{2mm}
\begin{tabularx}{\linewidth}{ccccc|cc|ccc}
\toprule
\multicolumn{5}{c|}{\textbf{Roles To Use}} & \multicolumn{2}{c|}{\textbf{ReXTime}} & \multicolumn{3}{c}{\textbf{Charades-STA}} \\
\cmidrule{1-5} \cmidrule{6-7} \cmidrule{8-10}
\textbf{A} & \textbf{G} & \textbf{V} & \textbf{P} & \textbf{G\%} & mIoU & Acc & R@0.5 & R@0.7 & mIoU \\
\midrule
\cmark &&&& 0\% & -- & 68.0 & -- & -- & -- \\
\cmark & \cmark &&& 100\% & 24.5 & 68.8 & -- & -- & -- \\
\cmark & \cmark & \cmark && 100\% & \underline{24.8} & 69.1 & -- & -- & -- \\
\cmark & \cmark & \cmark & \cmark & 100\% & 24.7 & \underline{69.2} & -- & -- & -- \\
\rowcolor{blue!7.5} \cmark & \cmark & \cmark & \cmark & 40\% & \textbf{26.7} & \textbf{70.0} & -- & -- & -- \\
\midrule
& \cmark &&&& -- & -- & \underline{47.2} & \underline{21.7} & \underline{42.0} \\
\rowcolor{blue!7.5} & \cmark & \cmark &&& -- & -- & \textbf{51.1} & \textbf{26.0} & \textbf{45.2} \\
\bottomrule
\end{tabularx}
\label{tab:role}
\end{minipage}
\vspace{-4mm}
\end{table}

\vspace{-1mm}
\subsection{Q2: The Advantages of Chain-of-LoRA}
\vspace{-1mm}

Table~\ref{tab:chain} studies the effect of role integration on VideoMind-2B. First, text-based CoT does not improve the base model, highlighting the need for a vision-centric reasoning strategy. Second, the key capabilities of roles may conflict with one another, thus only sub-optimal performance can be achieved via joint training. Compared to the all-distributed approach that requires multiple copies (4$\times$) of weights, Chain-of-LoRA offers the best balance between effectiveness and efficiency.

\vspace{-1mm}
\subsection{Q3: Key Ablation Studies}
\vspace{-1mm}

\paragraph{Effect of Individual Roles} The contributions of different roles are studied in Table~\ref{tab:role}. Our observations are as follows: (1) \texttt{\textbf{Grounder:}} By identifying visual cues, the grounder can slightly improve QA accuracy, indicating that the grounder is especially effective on long videos. (2) \texttt{\textbf{Verifier:}} Selecting the best candidate through the verifier improves grounding performance, yielding a consistent gain of 3.2 mIoU on Charades-STA. (3) \texttt{\textbf{Planner:}} Coordinating roles via the planner -- even when performing grounding on only 40\% samples (the remaining 60\% are directly processed by the answerer) -- boosts the accuracy from 69.2 to 70.0. This highlights the model's flexibility to adaptively determine whether to perform grounding under different temporal contexts.

\vspace{-1mm}
\subsection{Visualization}
\vspace{-1mm}

In Figure~\ref{fig:vis0}, we illustrate how VideoMind applies all roles to progressively derive the correct answer while avoiding potential mistakes. The planner determines what roles are needed, then calls the grounder to generate candidate moments. The verifier selects the most relevant segment (highlighted in yellow), which is then zoomed-in and passed to the answerer for further reasoning.

\vspace{-1mm}
\section{Conclusion}
\vspace{-1mm}

In this work, we introduced \textbf{VideoMind}, a video-language agent designed for temporal-grounded video reasoning. Our approach employs an agentic workflow consisting of four carefully designed roles along with a \textbf{Chain-of-LoRA} strategy to flexibly switch among them. Extensive experiments on Grounded VideoQA, Video Temporal Grounding, and General VideoQA tasks demonstrate the effectiveness and significance of our method, particularly in long-form video reasoning by providing evidence-based answers. We hope this work inspires future advancements in agentic reasoning.

\paragraph{Limitations \& Future Work} We acknowledge that our method requires careful optimization of individual designs and preparation of training data. In our future work, we will investigate (1) the possibility of joint-optimization of multiple roles and (2) the integration of audio modality.

\clearpage

\section*{Acknowledgements}

This research is supported by The Hong Kong Research Grants Council (GRF-15229423) and by Tencent's research donation. We also acknowledge The University Research Facility in Big Data Analytics (UBDA) at The Hong Kong Polytechnic University for providing computing resources that have contributed to the research results reported within this paper. Mike Shou does not receive any funding for this work.

\section*{Ethics Statement}

This study focuses on algorithmic innovations for improving the visual reasoning capabilities of large multi-modal models. It does not involve human subjects, private data, or any potentially harmful insights. All datasets used are publicly available and widely adopted in the community. We acknowledge the potential risks of misuse associated with LLMs and LMMs, including the bias propagation and harmful content generation. However, this study does not directly address the deployment or generation. Instead, it contributes to the understanding of the model architecture and the reasoning mechanism. To the best of our knowledge, our research complies with the ICLR Code of Ethics and does not involve any known violations or harms.

\section*{Reproducibility Statement}

We are committed to ensuring the full reproducibility of this study. To achieve this, we have provided key hyperparameters settings in Section~\ref{sec:method}, formulation of inference pipeline in Section~\ref{sec:inference}, implementation details in Section~\ref{sec:implementation}, evaluation metrics in Section~\ref{sec:benchmarks}, and prompt templates in Section~\ref{sec:prompt}. We also open-source all the code, model checkpoints, data, and training logs in this study to facilitate future research in this direction.

\bibliographystyle{main}
\bibliography{main}

\begin{thebibliography}{115}
\providecommand{\natexlab}[1]{#1}
\providecommand{\url}[1]{\texttt{#1}}
\expandafter\ifx\csname urlstyle\endcsname\relax
  \providecommand{\doi}[1]{doi: #1}\else
  \providecommand{\doi}{doi: \begingroup \urlstyle{rm}\Url}\fi

\bibitem[Abdin et~al.(2024)Abdin, Jacobs, Awan, Aneja, Awadallah, Awadalla, Bach, Bahree, Bakhtiari, Behl, et~al.]{phi3}
Marah Abdin, Sam~Ade Jacobs, Ammar~Ahmad Awan, Jyoti Aneja, Ahmed Awadallah, Hany Awadalla, Nguyen Bach, Amit Bahree, Arash Bakhtiari, Harkirat Behl, et~al.
\newblock Phi-3 technical report: A highly capable language model locally on your phone.
\newblock \emph{arXiv:2404.14219}, 2024.

\bibitem[Achiam et~al.(2023)Achiam, Adler, Agarwal, Ahmad, Akkaya, Aleman, Almeida, Altenschmidt, Altman, Anadkat, et~al.]{gpt4}
Josh Achiam, Steven Adler, Sandhini Agarwal, Lama Ahmad, Ilge Akkaya, Florencia~Leoni Aleman, Diogo Almeida, Janko Altenschmidt, Sam Altman, Shyamal Anadkat, et~al.
\newblock Gpt-4 technical report.
\newblock \emph{arXiv:2303.08774}, 2023.

\bibitem[Anthropic(2025)]{claude3.5sonnet}
Anthropic.
\newblock Claude 3.5 sonnet model card, 2025.
\newblock URL \url{https://www.anthropic.com/news/claude-3-5-sonnet}.

\bibitem[Bai et~al.(2025)Bai, Chen, Liu, Wang, Ge, Song, Dang, Wang, Wang, Tang, et~al.]{qwen2.5vl}
Shuai Bai, Keqin Chen, Xuejing Liu, Jialin Wang, Wenbin Ge, Sibo Song, Kai Dang, Peng Wang, Shijie Wang, Jun Tang, et~al.
\newblock Qwen2.5-vl technical report.
\newblock \emph{arXiv:2502.13923}, 2025.

\bibitem[Chakraborty et~al.(2024)Chakraborty, Ghosal, Yin, Manocha, Wang, Bedi, and Huang]{transferqstar}
Souradip Chakraborty, Soumya~Suvra Ghosal, Ming Yin, Dinesh Manocha, Mengdi Wang, Amrit~Singh Bedi, and Furong Huang.
\newblock Transfer q star: Principled decoding for llm alignment.
\newblock \emph{arXiv:2405.20495}, 2024.

\bibitem[Chen et~al.(2024{\natexlab{a}})Chen, Liu, Huang, He, Pei, Xu, Wang, Lu, and Wang]{cgbench}
Guo Chen, Yicheng Liu, Yifei Huang, Yuping He, Baoqi Pei, Jilan Xu, Yali Wang, Tong Lu, and Limin Wang.
\newblock Cg-bench: Clue-grounded question answering benchmark for long video understanding.
\newblock \emph{arXiv:2412.12075}, 2024{\natexlab{a}}.

\bibitem[Chen et~al.(2024{\natexlab{b}})Chen, Lv, Wu, Lin, Song, Gao, Liu, Gao, Mao, and Shou]{videollmol}
Joya Chen, Zhaoyang Lv, Shiwei Wu, Kevin~Qinghong Lin, Chenan Song, Difei Gao, Jia-Wei Liu, Ziteng Gao, Dongxing Mao, and Mike~Zheng Shou.
\newblock Videollm-online: Online video large language model for streaming video.
\newblock In \emph{CVPR}, pp.\  18407--18418, 2024{\natexlab{b}}.

\bibitem[Chen et~al.(2024{\natexlab{c}})Chen, Liao, Lin, Yu, Chen, and Wang]{rextime}
Jr-Jen Chen, Yu-Chien Liao, Hsi-Che Lin, Yu-Chu Yu, Yen-Chun Chen, and Frank Wang.
\newblock Rextime: A benchmark suite for reasoning-across-time in videos.
\newblock In \emph{NeurIPS}, pp.\  28662--28673, 2024{\natexlab{c}}.

\bibitem[Chen et~al.(2024{\natexlab{d}})Chen, Wei, Li, Dong, Zhang, Zang, Chen, Duan, Tang, Yuan, et~al.]{sharegpt4video}
Lin Chen, Xilin Wei, Jinsong Li, Xiaoyi Dong, Pan Zhang, Yuhang Zang, Zehui Chen, Haodong Duan, Zhenyu Tang, Li~Yuan, et~al.
\newblock Sharegpt4video: Improving video understanding and generation with better captions.
\newblock In \emph{NeurIPS}, pp.\  19472--19495, 2024{\natexlab{d}}.

\bibitem[Chen et~al.(2025{\natexlab{a}})Chen, Di, and Xie]{multihopegoqa}
Qirui Chen, Shangzhe Di, and Weidi Xie.
\newblock Grounded multi-hop videoqa in long-form egocentric videos.
\newblock In \emph{AAAI}, pp.\  2159--2167, 2025{\natexlab{a}}.

\bibitem[Chen et~al.(2024{\natexlab{e}})Chen, Lan, Yuan, Jie, and Ma]{timemarker}
Shimin Chen, Xiaohan Lan, Yitian Yuan, Zequn Jie, and Lin Ma.
\newblock Timemarker: A versatile video-llm for long and short video understanding with superior temporal localization ability.
\newblock \emph{arXiv:2411.18211}, 2024{\natexlab{e}}.

\bibitem[Chen et~al.(2025{\natexlab{b}})Chen, Huang, Shi, Hu, Ye, Zhu, Liu, Molchanov, Kautz, Qi, et~al.]{longvilar1}
Yukang Chen, Wei Huang, Baifeng Shi, Qinghao Hu, Hanrong Ye, Ligeng Zhu, Zhijian Liu, Pavlo Molchanov, Jan Kautz, Xiaojuan Qi, et~al.
\newblock Scaling rl to long videos.
\newblock \emph{arXiv:2507.07966}, 2025{\natexlab{b}}.

\bibitem[DeepMind(2025)]{gemini2.5}
Google DeepMind.
\newblock Gemini 2.5: Our most intelligent ai model, 2025.
\newblock URL \url{https://blog.google/technology/google-deepmind/gemini-model-thinking-updates-march-2025/}.

\bibitem[Fan et~al.(2024)Fan, Ma, Wu, Du, Li, Gao, and Li]{videoagentmem}
Yue Fan, Xiaojian Ma, Rujie Wu, Yuntao Du, Jiaqi Li, Zhi Gao, and Qing Li.
\newblock Videoagent: A memory-augmented multimodal agent for video understanding.
\newblock \emph{arXiv:2403.11481}, 2024.

\bibitem[Fei et~al.(2024)Fei, Li, Deng, Wang, Liu, and Wang]{videoccam}
Jiajun Fei, Dian Li, Zhidong Deng, Zekun Wang, Gang Liu, and Hui Wang.
\newblock Video-ccam: Enhancing video-language understanding with causal cross-attention masks for short and long videos.
\newblock \emph{arXiv:2408.14023}, 2024.

\bibitem[Feng et~al.(2025)Feng, Gong, Li, Guo, Wang, Peng, Wu, Zhang, Wang, and Yue]{videor1}
Kaituo Feng, Kaixiong Gong, Bohao Li, Zonghao Guo, Yibing Wang, Tianshuo Peng, Junfei Wu, Xiaoying Zhang, Benyou Wang, and Xiangyu Yue.
\newblock Video-r1: Reinforcing video reasoning in mllms.
\newblock \emph{arXiv:2503.21776}, 2025.

\bibitem[Fu et~al.(2024{\natexlab{a}})Fu, Dai, Luo, Li, Ren, Zhang, Wang, Zhou, Shen, Zhang, et~al.]{videomme}
Chaoyou Fu, Yuhan Dai, Yongdong Luo, Lei Li, Shuhuai Ren, Renrui Zhang, Zihan Wang, Chenyu Zhou, Yunhang Shen, Mengdan Zhang, et~al.
\newblock Video-mme: The first-ever comprehensive evaluation benchmark of multi-modal llms in video analysis.
\newblock \emph{arXiv:2405.21075}, 2024{\natexlab{a}}.

\bibitem[Fu et~al.(2024{\natexlab{b}})Fu, Lin, Long, Shen, Dai, Zhao, Zhang, Dong, Li, Wang, et~al.]{vita}
Chaoyou Fu, Haojia Lin, Zuwei Long, Yunhang Shen, Yuhang Dai, Meng Zhao, Yi-Fan Zhang, Shaoqi Dong, Yangze Li, Xiong Wang, et~al.
\newblock Vita: Towards open-source interactive omni multimodal llm.
\newblock \emph{arXiv:2408.05211}, 2024{\natexlab{b}}.

\bibitem[Gao et~al.(2023)Gao, Ji, Zhou, Lin, Chen, Fan, and Shou]{assistgpt}
Difei Gao, Lei Ji, Luowei Zhou, Kevin~Qinghong Lin, Joya Chen, Zihan Fan, and Mike~Zheng Shou.
\newblock Assistgpt: A general multi-modal assistant that can plan, execute, inspect, and learn.
\newblock \emph{arXiv:2306.08640}, 2023.

\bibitem[Gao et~al.(2017)Gao, Sun, Yang, and Nevatia]{charadessta}
Jiyang Gao, Chen Sun, Zhenheng Yang, and Ram Nevatia.
\newblock Tall: Temporal activity localization via language query.
\newblock In \emph{ICCV}, pp.\  5267--5275, 2017.

\bibitem[Grauman et~al.(2022)Grauman, Westbury, Byrne, Chavis, Furnari, Girdhar, Hamburger, Jiang, Liu, Liu, et~al.]{ego4d}
Kristen Grauman, Andrew Westbury, Eugene Byrne, Zachary Chavis, Antonino Furnari, Rohit Girdhar, Jackson Hamburger, Hao Jiang, Miao Liu, Xingyu Liu, et~al.
\newblock Ego4d: Around the world in 3,000 hours of egocentric video.
\newblock In \emph{CVPR}, pp.\  18995--19012, 2022.

\bibitem[Guo et~al.(2025)Guo, Yang, Zhang, Song, Zhang, Xu, Zhu, Ma, Wang, Bi, et~al.]{deepseekr1}
Daya Guo, Dejian Yang, Haowei Zhang, Junxiao Song, Ruoyu Zhang, Runxin Xu, Qihao Zhu, Shirong Ma, Peiyi Wang, Xiao Bi, et~al.
\newblock Deepseek-r1: Incentivizing reasoning capability in llms via reinforcement learning.
\newblock \emph{arXiv:2501.12948}, 2025.

\bibitem[Guo et~al.(2024)Guo, Liu, Li, Liu, Chen, and Tang]{trace}
Yongxin Guo, Jingyu Liu, Mingda Li, Qingbin Liu, Xi~Chen, and Xiaoying Tang.
\newblock Trace: Temporal grounding video llm via causal event modeling.
\newblock \emph{arXiv:2410.05643}, 2024.

\bibitem[Hu et~al.(2022)Hu, Shen, Wallis, Allen-Zhu, Li, Wang, Wang, Chen, et~al.]{lora}
Edward~J Hu, Yelong Shen, Phillip Wallis, Zeyuan Allen-Zhu, Yuanzhi Li, Shean Wang, Lu~Wang, Weizhu Chen, et~al.
\newblock Lora: Low-rank adaptation of large language models.
\newblock \emph{ICLR}, 2022.

\bibitem[Huang et~al.(2024{\natexlab{a}})Huang, Wang, Chen, Song, and Zhu]{vtimellm}
Bin Huang, Xin Wang, Hong Chen, Zihan Song, and Wenwu Zhu.
\newblock Vtimellm: Empower llm to grasp video moments.
\newblock In \emph{CVPR}, pp.\  14271--14280, 2024{\natexlab{a}}.

\bibitem[Huang et~al.(2024{\natexlab{b}})Huang, Liao, Radhakrishnan, Yin, Molchanov, Yu, and Kautz]{lita}
De-An Huang, Shijia Liao, Subhashree Radhakrishnan, Hongxu Yin, Pavlo Molchanov, Zhiding Yu, and Jan Kautz.
\newblock Lita: Language instructed temporal-localization assistant.
\newblock \emph{arXiv:2403.19046}, 2024{\natexlab{b}}.

\bibitem[Jiang et~al.(2024)Jiang, He, Zeng, Wei, Ku, Liu, and Chen]{mantis}
Dongfu Jiang, Xuan He, Huaye Zeng, Cong Wei, Max Ku, Qian Liu, and Wenhu Chen.
\newblock Mantis: Interleaved multi-image instruction tuning.
\newblock \emph{arXiv:2405.01483}, 2024.

\bibitem[Jin et~al.(2023)Jin, Takanobu, Zhang, Cao, and Yuan]{chatunivi}
Peng Jin, Ryuichi Takanobu, Caiwan Zhang, Xiaochun Cao, and Li~Yuan.
\newblock Chat-univi: Unified visual representation empowers large language models with image and video understanding.
\newblock \emph{arXiv:2311.08046}, 2023.

\bibitem[Kahatapitiya et~al.(2024)Kahatapitiya, Ranasinghe, Park, and Ryoo]{langrepo}
Kumara Kahatapitiya, Kanchana Ranasinghe, Jongwoo Park, and Michael~S Ryoo.
\newblock Language repository for long video understanding.
\newblock \emph{arXiv:2403.14622}, 2024.

\bibitem[Krishna et~al.(2017)Krishna, Hata, Ren, Fei-Fei, and Carlos~Niebles]{activitynetcaptions}
Ranjay Krishna, Kenji Hata, Frederic Ren, Li~Fei-Fei, and Juan Carlos~Niebles.
\newblock Dense-captioning events in videos.
\newblock In \emph{ICCV}, pp.\  706--715, 2017.

\bibitem[Laurencon et~al.(2024)Laurencon, Tronchon, Cord, and Sanh]{idefics2}
Hugo Laurencon, Leo Tronchon, Matthieu Cord, and Victor Sanh.
\newblock What matters when building vision-language models?
\newblock In \emph{NeurIPS}, pp.\  87874--87907, 2024.

\bibitem[Lei et~al.(2020)Lei, Yu, Berg, and Bansal]{tvr}
Jie Lei, Licheng Yu, Tamara~L Berg, and Mohit Bansal.
\newblock Tvr: A large-scale dataset for video-subtitle moment retrieval.
\newblock In \emph{ECCV}, pp.\  447--463, 2020.

\bibitem[Lei et~al.(2021)Lei, Berg, and Bansal]{qvhighlights}
Jie Lei, Tamara~L Berg, and Mohit Bansal.
\newblock Qvhighlights: Detecting moments and highlights in videos via natural language queries.
\newblock In \emph{NeurIPS}, 2021.

\bibitem[Li et~al.(2024{\natexlab{a}})Li, Zhang, Guo, Zhang, Li, Zhang, Zhang, Zhang, Li, Liu, et~al.]{llavaonevision}
Bo~Li, Yuanhan Zhang, Dong Guo, Renrui Zhang, Feng Li, Hao Zhang, Kaichen Zhang, Peiyuan Zhang, Yanwei Li, Ziwei Liu, et~al.
\newblock Llava-onevision: Easy visual task transfer.
\newblock \emph{arXiv:2408.03326}, 2024{\natexlab{a}}.

\bibitem[Li et~al.(2025{\natexlab{a}})Li, Chen, Wei, Huang, Hui, Gao, Wei, and Liu]{llavast}
Hongyu Li, Jinyu Chen, Ziyu Wei, Shaofei Huang, Tianrui Hui, Jialin Gao, Xiaoming Wei, and Si~Liu.
\newblock Llava-st: A multimodal large language model for fine-grained spatial-temporal understanding.
\newblock In \emph{CVPR}, pp.\  8592--8603, 2025{\natexlab{a}}.

\bibitem[Li et~al.(2023{\natexlab{a}})Li, He, Wang, Li, Wang, Luo, Wang, Wang, and Qiao]{videochat}
Kunchang Li, Yinan He, Yi~Wang, Yizhuo Li, Wenhai Wang, Ping Luo, Yali Wang, Limin Wang, and Yu~Qiao.
\newblock Videochat: Chat-centric video understanding.
\newblock \emph{arXiv:2305.06355}, 2023{\natexlab{a}}.

\bibitem[Li et~al.(2024{\natexlab{b}})Li, Wang, He, Li, Wang, Liu, Wang, Xu, Chen, Luo, et~al.]{mvbench}
Kunchang Li, Yali Wang, Yinan He, Yizhuo Li, Yi~Wang, Yi~Liu, Zun Wang, Jilan Xu, Guo Chen, Ping Luo, et~al.
\newblock Mvbench: A comprehensive multi-modal video understanding benchmark.
\newblock In \emph{CVPR}, pp.\  22195--22206, 2024{\natexlab{b}}.

\bibitem[Li et~al.(2023{\natexlab{b}})Li, Xie, Xie, Zhao, Zhang, Zheng, Zhao, and Zhang]{momentdiff}
Pandeng Li, Chen-Wei Xie, Hongtao Xie, Liming Zhao, Lei Zhang, Yun Zheng, Deli Zhao, and Yongdong Zhang.
\newblock Momentdiff: Generative video moment retrieval from random to real.
\newblock In \emph{NeurIPS}, pp.\  65948--65966, 2023{\natexlab{b}}.

\bibitem[Li et~al.(2025{\natexlab{b}})Li, Di, Zhai, Huang, Wang, and Xie]{unitime}
Zeqian Li, Shangzhe Di, Zhonghua Zhai, Weilin Huang, Yanfeng Wang, and Weidi Xie.
\newblock Universal video temporal grounding with generative multi-modal large language models.
\newblock \emph{arXiv:2506.18883}, 2025{\natexlab{b}}.

\bibitem[Lightman et~al.(2023)Lightman, Kosaraju, Burda, Edwards, Baker, Lee, Leike, Schulman, Sutskever, and Cobbe]{letsverify}
Hunter Lightman, Vineet Kosaraju, Yuri Burda, Harrison Edwards, Bowen Baker, Teddy Lee, Jan Leike, John Schulman, Ilya Sutskever, and Karl Cobbe.
\newblock Let's verify step by step.
\newblock In \emph{ICLR}, 2023.

\bibitem[Lin et~al.(2023{\natexlab{a}})Lin, Zhu, Ye, Ning, Jin, and Yuan]{videollava}
Bin Lin, Bin Zhu, Yang Ye, Munan Ning, Peng Jin, and Li~Yuan.
\newblock Video-llava: Learning united visual representation by alignment before projection.
\newblock \emph{arXiv:2311.10122}, 2023{\natexlab{a}}.

\bibitem[Lin et~al.(2024{\natexlab{a}})Lin, Yin, Ping, Molchanov, Shoeybi, and Han]{vila}
Ji~Lin, Hongxu Yin, Wei Ping, Pavlo Molchanov, Mohammad Shoeybi, and Song Han.
\newblock Vila: On pre-training for visual language models.
\newblock In \emph{CVPR}, pp.\  26689--26699, 2024{\natexlab{a}}.

\bibitem[Lin \& Shou(2025)Lin and Shou]{vlog}
Kevin~Qinghong Lin and Mike~Zheng Shou.
\newblock Vlog: Video-language models by generative retrieval of narration vocabulary.
\newblock In \emph{CVPR}, pp.\  3218--3228, 2025.

\bibitem[Lin et~al.(2022)Lin, Wang, Soldan, Wray, Yan, Xu, Gao, Tu, Zhao, Kong, et~al.]{egovlp}
Kevin~Qinghong Lin, Jinpeng Wang, Mattia Soldan, Michael Wray, Rui Yan, Eric~Z Xu, Difei Gao, Rong-Cheng Tu, Wenzhe Zhao, Weijie Kong, et~al.
\newblock Egocentric video-language pretraining.
\newblock In \emph{NeurIPS}, pp.\  7575--7586, 2022.

\bibitem[Lin et~al.(2023{\natexlab{b}})Lin, Zhang, Chen, Pramanick, Gao, Wang, Yan, and Shou]{univtg}
Kevin~Qinghong Lin, Pengchuan Zhang, Joya Chen, Shraman Pramanick, Difei Gao, Alex~Jinpeng Wang, Rui Yan, and Mike~Zheng Shou.
\newblock Univtg: Towards unified video-language temporal grounding.
\newblock In \emph{CVPR}, pp.\  2794--2804, 2023{\natexlab{b}}.

\bibitem[Lin et~al.(2024{\natexlab{b}})Lin, Zhang, Gao, Xia, Chen, Gao, Xie, Xiao, and Shou]{movieseq}
Kevin~Qinghong Lin, Pengchuan Zhang, Difei Gao, Xide Xia, Joya Chen, Ziteng Gao, Jinheng Xie, Xuhong Xiao, and Mike~Zheng Shou.
\newblock Learning video context as interleaved multimodal sequences.
\newblock In \emph{ECCV}, pp.\  375--396, 2024{\natexlab{b}}.

\bibitem[Lin et~al.(2017)Lin, Goyal, Girshick, He, and Dollár]{retinenet}
Tsung-Yi Lin, Priya Goyal, Ross Girshick, Kaiming He, and Piotr Dollár.
\newblock Focal loss for dense object detection.
\newblock In \emph{ICCV}, pp.\  2980--2988, 2017.

\bibitem[Liu et~al.(2023)Liu, Li, Wu, and Lee]{llava}
Haotian Liu, Chunyuan Li, Qingyang Wu, and Yong~Jae Lee.
\newblock Visual instruction tuning.
\newblock In \emph{NeurIPS}, pp.\  34892--34916, 2023.

\bibitem[Liu et~al.(2024{\natexlab{a}})Liu, Li, Li, Li, Zhang, Shen, and Lee]{llavanext}
Haotian Liu, Chunyuan Li, Yuheng Li, Bo~Li, Yuanhan Zhang, Sheng Shen, and Yong~Jae Lee.
\newblock Llava-next: Improved reasoning, ocr, and world knowledge, 2024{\natexlab{a}}.
\newblock URL \url{https://llava-vl.github.io/blog/2024-01-30-llava-next/}.

\bibitem[Liu et~al.(2024{\natexlab{b}})Liu, Wang, Ma, Wu, Ma, Wei, Jiao, Wu, and Hu]{kangaroo}
Jiajun Liu, Yibing Wang, Hanghang Ma, Xiaoping Wu, Xiaoqi Ma, Xiaoming Wei, Jianbin Jiao, Enhua Wu, and Jie Hu.
\newblock Kangaroo: A powerful video-language model supporting long-context video input.
\newblock \emph{arXiv:2408.15542}, 2024{\natexlab{b}}.

\bibitem[Liu et~al.(2024{\natexlab{c}})Liu, Li, Tang, Ge, Shan, and Li]{stllm}
Ruyang Liu, Chen Li, Haoran Tang, Yixiao Ge, Ying Shan, and Ge~Li.
\newblock St-llm: Large language models are effective temporal learners.
\newblock In \emph{ECCV}, pp.\  1--18, 2024{\natexlab{c}}.

\bibitem[Liu et~al.(2022)Liu, Li, Wu, Chen, Shan, and Qie]{umt}
Ye~Liu, Siyuan Li, Yang Wu, Chang~Wen Chen, Ying Shan, and Xiaohu Qie.
\newblock Umt: Unified multi-modal transformers for joint video moment retrieval and highlight detection.
\newblock In \emph{CVPR}, pp.\  3042--3051, 2022.

\bibitem[Liu et~al.(2024{\natexlab{d}})Liu, He, Li, Kim, Wei, Pfister, and Chen]{r2tuning}
Ye~Liu, Jixuan He, Wanhua Li, Junsik Kim, Donglai Wei, Hanspeter Pfister, and Chang~Wen Chen.
\newblock $r^2$-tuning: Efficient image-to-video transfer learning for video temporal grounding.
\newblock In \emph{ECCV}, 2024{\natexlab{d}}.

\bibitem[Liu et~al.(2024{\natexlab{e}})Liu, Ma, Qi, Wu, Shan, and Chen]{etbench}
Ye~Liu, Zongyang Ma, Zhongang Qi, Yang Wu, Ying Shan, and Chang~W Chen.
\newblock E.t. bench: Towards open-ended event-level video-language understanding.
\newblock In \emph{NeurIPS}, pp.\  32076--32110, 2024{\natexlab{e}}.

\bibitem[Liu et~al.(2025)Liu, Ma, Pu, Qi, Wu, Ying, and Chen]{unipixel}
Ye~Liu, Zongyang Ma, Junfu Pu, Zhongang Qi, Yang Wu, Shan Ying, and Chang~Wen Chen.
\newblock Unipixel: Unified object referring and segmentation for pixel-level visual reasoning.
\newblock In \emph{NeurIPS}, 2025.

\bibitem[Loshchilov \& Hutter(2019)Loshchilov and Hutter]{adamw}
Ilya Loshchilov and Frank Hutter.
\newblock Decoupled weight decay regularization.
\newblock In \emph{ICLR}, 2019.

\bibitem[Lu et~al.(2022)Lu, Mishra, Xia, Qiu, Chang, Zhu, Tafjord, Clark, and Kalyan]{scienceqa}
Pan Lu, Swaroop Mishra, Tanglin Xia, Liang Qiu, Kai-Wei Chang, Song-Chun Zhu, Oyvind Tafjord, Peter Clark, and Ashwin Kalyan.
\newblock Learn to explain: Multimodal reasoning via thought chains for science question answering.
\newblock In \emph{NeurIPS}, pp.\  2507--2521, 2022.

\bibitem[Ma et~al.(2025)Ma, Chen, Zhang, Qi, Yuan, Zhu, Zhuo, Li, Liu, Li, Shan, and Hu]{visionmath}
Zongyang Ma, Yuxin Chen, Ziqi Zhang, Zhongang Qi, Chunfeng Yuan, Shaojie Zhu, Chengxiang Zhuo, Bing Li, Ye~Liu, Zang Li, Ying Shan, and Weiming Hu.
\newblock Visionmath: Vision-form mathematical problem-solving.
\newblock In \emph{ICCV}, 2025.

\bibitem[Maaz et~al.(2023)Maaz, Rasheed, Khan, and Khan]{videochatgpt}
Muhammad Maaz, Hanoona Rasheed, Salman Khan, and Fahad~Shahbaz Khan.
\newblock Video-chatgpt: Towards detailed video understanding via large vision and language models.
\newblock \emph{arXiv:2306.05424}, 2023.

\bibitem[Maaz et~al.(2024)Maaz, Rasheed, Khan, and Khan]{videogpt}
Muhammad Maaz, Hanoona Rasheed, Salman Khan, and Fahad Khan.
\newblock Videogpt+: Integrating image and video encoders for enhanced video understanding.
\newblock \emph{arXiv:2406.09418}, 2024.

\bibitem[Moon et~al.(2023)Moon, Hyun, Park, Park, and Heo]{qddetr}
WonJun Moon, Sangeek Hyun, SangUk Park, Dongchan Park, and Jae-Pil Heo.
\newblock Query-dependent video representation for moment retrieval and highlight detection.
\newblock In \emph{CVPR}, pp.\  23023--23033, 2023.

\bibitem[OpenAI(2023)]{gpt4v}
OpenAI.
\newblock Gpt-4v(ision) system card, 2023.
\newblock URL \url{https://openai.com/index/gpt-4v-system-card/}.

\bibitem[OpenAI(2024{\natexlab{a}})]{gpt4o}
OpenAI.
\newblock Gpt-4o system card, 2024{\natexlab{a}}.
\newblock URL \url{https://openai.com/index/gpt-4o-system-card/}.

\bibitem[OpenAI(2024{\natexlab{b}})]{openaio1}
OpenAI.
\newblock Openai o1 system card, 2024{\natexlab{b}}.
\newblock URL \url{https://openai.com/index/openai-o1-system-card/}.

\bibitem[OpenAI(2025)]{gpt5}
OpenAI.
\newblock Gpt-5 system card, 2025.
\newblock URL \url{https://openai.com/index/gpt-5-system-card/}.

\bibitem[OpenGVLab(2024)]{internvl2}
OpenGVLab.
\newblock Internvl2: Better than the best—expanding performance boundaries of open-source multimodal models with the progressive scaling strategy, 2024.
\newblock URL \url{https://internvl.github.io/blog/2024-07-02-InternVL-2.0/}.

\bibitem[Qian et~al.(2024{\natexlab{a}})Qian, Li, Wu, Ye, Fei, Chua, Zhuang, and Tang]{momentor}
Long Qian, Juncheng Li, Yu~Wu, Yaobo Ye, Hao Fei, Tat-Seng Chua, Yueting Zhuang, and Siliang Tang.
\newblock Momentor: Advancing video large language model with fine-grained temporal reasoning.
\newblock \emph{arXiv:2402.11435}, 2024{\natexlab{a}}.

\bibitem[Qian et~al.(2024{\natexlab{b}})Qian, Dong, Zhang, Zang, Ding, Lin, and Wang]{videostreaming}
Rui Qian, Xiaoyi Dong, Pan Zhang, Yuhang Zang, Shuangrui Ding, Dahua Lin, and Jiaqi Wang.
\newblock Streaming long video understanding with large language models.
\newblock In \emph{NeurIPS}, pp.\  119336--119360, 2024{\natexlab{b}}.

\bibitem[Qu et~al.(2024)Qu, Chen, Liu, Li, and Zhao]{chatvtg}
Mengxue Qu, Xiaodong Chen, Wu~Liu, Alicia Li, and Yao Zhao.
\newblock Chatvtg: Video temporal grounding via chat with video dialogue large language models.
\newblock In \emph{CVPR}, pp.\  1847--1856, 2024.

\bibitem[Ramakrishnan et~al.(2023)Ramakrishnan, Al-Halah, and Grauman]{ego4dnaq}
Santhosh~Kumar Ramakrishnan, Ziad Al-Halah, and Kristen Grauman.
\newblock Naq: Leveraging narrations as queries to supervise episodic memory.
\newblock In \emph{CVPR}, pp.\  6694--6703, 2023.

\bibitem[Regneri et~al.(2013)Regneri, Rohrbach, Wetzel, Thater, Schiele, and Pinkal]{tacos}
Michaela Regneri, Marcus Rohrbach, Dominikus Wetzel, Stefan Thater, Bernt Schiele, and Manfred Pinkal.
\newblock Grounding action descriptions in videos.
\newblock \emph{Transactions of the Association for Computational Linguistics}, 1:\penalty0 25--36, 2013.

\bibitem[Reid et~al.(2024)Reid, Savinov, Teplyashin, Lepikhin, Lillicrap, Alayrac, Soricut, Lazaridou, Firat, Schrittwieser, et~al.]{gemini1.5}
Machel Reid, Nikolay Savinov, Denis Teplyashin, Dmitry Lepikhin, Timothy Lillicrap, Jean-baptiste Alayrac, Radu Soricut, Angeliki Lazaridou, Orhan Firat, Julian Schrittwieser, et~al.
\newblock Gemini 1.5: Unlocking multimodal understanding across millions of tokens of context.
\newblock \emph{arXiv:2403.05530}, 2024.

\bibitem[Ren et~al.(2024)Ren, Yao, Li, Sun, and Hou]{timechat}
Shuhuai Ren, Linli Yao, Shicheng Li, Xu~Sun, and Lu~Hou.
\newblock Timechat: A time-sensitive multimodal large language model for long video understanding.
\newblock In \emph{CVPR}, pp.\  14313--14323, 2024.

\bibitem[Seed(2026)]{seed2.0}
ByteDance Seed.
\newblock Seed2.0 model card: Towards intelligence frontier for real-world complexity, 2026.
\newblock URL \url{https://seed.bytedance.com/en/seed2}.

\bibitem[Shinn et~al.(2023)Shinn, Cassano, Gopinath, Narasimhan, and Yao]{reflexion}
Noah Shinn, Federico Cassano, Ashwin Gopinath, Karthik Narasimhan, and Shunyu Yao.
\newblock Reflexion: Language agents with verbal reinforcement learning.
\newblock In \emph{NeurIPS}, pp.\  8634--8652, 2023.

\bibitem[Song et~al.(2023)Song, Chai, Wang, Zhang, Zhou, Wu, Guo, Ye, Lu, Hwang, et~al.]{moviechat}
Enxin Song, Wenhao Chai, Guanhong Wang, Yucheng Zhang, Haoyang Zhou, Feiyang Wu, Xun Guo, Tian Ye, Yan Lu, Jenq-Neng Hwang, et~al.
\newblock Moviechat: From dense token to sparse memory for long video understanding.
\newblock \emph{arXiv:2307.16449}, 2023.

\bibitem[Suris et~al.(2023)Suris, Menon, and Vondrick]{vipergpt}
Didac Suris, Sachit Menon, and Carl Vondrick.
\newblock Vipergpt: Visual inference via python execution for reasoning.
\newblock In \emph{ICCV}, pp.\  11888--11898, 2023.

\bibitem[Thawakar et~al.(2025)Thawakar, Dissanayake, More, Thawkar, Heakl, Ahsan, Li, Zumri, Lahoud, Anwer, et~al.]{llamavo1}
Omkar Thawakar, Dinura Dissanayake, Ketan More, Ritesh Thawkar, Ahmed Heakl, Noor Ahsan, Yuhao Li, Mohammed Zumri, Jean Lahoud, Rao~Muhammad Anwer, et~al.
\newblock Llamav-o1: Rethinking step-by-step visual reasoning in llms.
\newblock \emph{arXiv:2501.06186}, 2025.

\bibitem[Vaswani et~al.(2017)Vaswani, Shazeer, Parmar, Uszkoreit, Jones, Gomez, Kaiser, and Polosukhin]{transformer}
Ashish Vaswani, Noam Shazeer, Niki Parmar, Jakob Uszkoreit, Llion Jones, Aidan~N Gomez, Lukasz Kaiser, and Illia Polosukhin.
\newblock Attention is all you need.
\newblock In \emph{NeurIPS}, pp.\  5998--6008, 2017.

\bibitem[Wang et~al.(2024{\natexlab{a}})Wang, Xu, Cheng, Diao, Zhou, Cao, Wang, Ge, and Huang]{groundedvideollm}
Haibo Wang, Zhiyang Xu, Yu~Cheng, Shizhe Diao, Yufan Zhou, Yixin Cao, Qifan Wang, Weifeng Ge, and Lifu Huang.
\newblock Grounded-videollm: Sharpening fine-grained temporal grounding in video large language models.
\newblock \emph{arXiv:2410.03290}, 2024{\natexlab{a}}.

\bibitem[Wang et~al.(2024{\natexlab{b}})Wang, Bai, Tan, Wang, Fan, Bai, Chen, Liu, Wang, Ge, et~al.]{qwen2vl}
Peng Wang, Shuai Bai, Sinan Tan, Shijie Wang, Zhihao Fan, Jinze Bai, Keqin Chen, Xuejing Liu, Jialin Wang, Wenbin Ge, et~al.
\newblock Qwen2-vl: Enhancing vision-language model's perception of the world at any resolution.
\newblock \emph{arXiv:2409.12191}, 2024{\natexlab{b}}.

\bibitem[Wang et~al.(2024{\natexlab{c}})Wang, He, Hong, Cheng, Zhang, Qi, Gu, Huang, Xu, Dong, et~al.]{lvbench}
Weihan Wang, Zehai He, Wenyi Hong, Yean Cheng, Xiaohan Zhang, Ji~Qi, Xiaotao Gu, Shiyu Huang, Bin Xu, Yuxiao Dong, et~al.
\newblock Lvbench: An extreme long video understanding benchmark.
\newblock \emph{arXiv:2406.08035}, 2024{\natexlab{c}}.

\bibitem[Wang et~al.(2024{\natexlab{d}})Wang, Zhang, Zohar, and Yeung-Levy]{videoagentlong}
Xiaohan Wang, Yuhui Zhang, Orr Zohar, and Serena Yeung-Levy.
\newblock Videoagent: Long-form video understanding with large language model as agent.
\newblock \emph{arXiv:2403.10517}, 2024{\natexlab{d}}.

\bibitem[Wang et~al.(2024{\natexlab{e}})Wang, Song, Tian, Yu, Peng, Mi, Huang, and Yu]{llmmcts}
Xiyao Wang, Linfeng Song, Ye~Tian, Dian Yu, Baolin Peng, Haitao Mi, Furong Huang, and Dong Yu.
\newblock Towards self-improvement of llms via mcts: Leveraging stepwise knowledge with curriculum preference learning.
\newblock \emph{arXiv:2410.06508}, 2024{\natexlab{e}}.

\bibitem[Wang et~al.(2024{\natexlab{f}})Wang, Meng, Liang, Wang, Liu, and Zhao]{hawkeye}
Yueqian Wang, Xiaojun Meng, Jianxin Liang, Yuxuan Wang, Qun Liu, and Dongyan Zhao.
\newblock Hawkeye: Training video-text llms for grounding text in videos.
\newblock \emph{arXiv:2403.10228}, 2024{\natexlab{f}}.

\bibitem[Wei et~al.(2022)Wei, Wang, Schuurmans, Bosma, Xia, Chi, Le, Zhou, et~al.]{cot}
Jason Wei, Xuezhi Wang, Dale Schuurmans, Maarten Bosma, Fei Xia, Ed~Chi, Quoc~V Le, Denny Zhou, et~al.
\newblock Chain-of-thought prompting elicits reasoning in large language models.
\newblock In \emph{NeurIPS}, pp.\  24824--24837, 2022.

\bibitem[Wen et~al.(2025)Wen, Xie, Lyu, Zhang, Wu, and Leong]{rsfmrieeg}
Junjian Wen, Linshan Xie, Qiuyi Lyu, Zhangjin Zhang, Ed~X Wu, and Alex~TL Leong.
\newblock Simultaneous rsfmri and eeg reveal the brain-wide actions of delta-gamma neural oscillations coupling.
\newblock In \emph{ISMRM}, 2025.

\bibitem[Wu et~al.(2024)Wu, Li, Chen, and Li]{longvideobench}
Haoning Wu, Dongxu Li, Bei Chen, and Junnan Li.
\newblock Longvideobench: A benchmark for long-context interleaved video-language understanding.
\newblock In \emph{NeurIPS}, pp.\  28828--28857, 2024.

\bibitem[Wu et~al.(2025)Wu, Liu, Liu, Liu, Nie, Lin, and Chen]{vtgsurvey}
Jianlong Wu, Wei Liu, Ye~Liu, Meng Liu, Liqiang Nie, Zhouchen Lin, and Chang~Wen Chen.
\newblock A survey on video temporal grounding with multimodal large language model.
\newblock \emph{arXiv:2508.10922}, 2025.

\bibitem[Xiao et~al.(2021)Xiao, Shang, Yao, and Chua]{nextqa}
Junbin Xiao, Xindi Shang, Angela Yao, and Tat-Seng Chua.
\newblock Next-qa: Next phase of question-answering to explaining temporal actions.
\newblock In \emph{CVPR}, pp.\  9777--9786, 2021.

\bibitem[Xiao et~al.(2024)Xiao, Yao, Li, and Chua]{nextgqa}
Junbin Xiao, Angela Yao, Yicong Li, and Tat-Seng Chua.
\newblock Can i trust your answer? visually grounded video question answering.
\newblock In \emph{CVPR}, pp.\  13204--13214, 2024.

\bibitem[Xie et~al.(2022)Xie, Wang, Ma, Chong, Lim, Cao, Khong, Wu, and Leong]{visualthalamus}
Linshan Xie, Xunda Wang, Teng Ma, Pit~Shan Chong, Lee~Wei Lim, Peng Cao, Pek-Lan Khong, Ed~X Wu, and Alex~TL Leong.
\newblock Are topographically segregated excitatory neurons in visual thalamus functionally diverse? an optogenetic fmri study.
\newblock In \emph{ISMRM}, 2022.

\bibitem[Xie et~al.(2023)Xie, Wang, Ma, Zeng, Wen, Cao, Wu, and Leong]{shortsinglepulse}
Linshan Xie, Xunda Wang, Teng Ma, Hang Zeng, Junjian Wen, Peng Cao, Ed~X Wu, and Alex~TL Leong.
\newblock Short single pulse optogenetic fmri mapping of downstream targets in thalamo-cortical pathways.
\newblock In \emph{ISMRM}, 2023.

\bibitem[Xie et~al.(2024{\natexlab{a}})Xie, Lin, Wang, Wen, Ma, Hu, Cao, Leong, and Wu]{eegfmri}
Linshan Xie, Xuehong Lin, Xunda Wang, Junjian Wen, Teng Ma, Jiahao Hu, Peng Cao, Alex~TL Leong, and Ed~X Wu.
\newblock Simultaneous eeg-fmri reveals spontaneous neural oscillatory activity in cingulate cortex underlying transient rsfmri network dynamics.
\newblock In \emph{ISMRM}, 2024{\natexlab{a}}.

\bibitem[Xie et~al.(2024{\natexlab{b}})Xie, Wang, Lin, Ma, Wen, Cao, Leong, and Wu]{singlepulse}
Linshan Xie, Xunda Wang, Xuehong Lin, Teng Ma, Junjian Wen, Peng Cao, Alex~TL Leong, and Ed~X Wu.
\newblock Single-pulse optogenetic perturbation of thalamo-cortical networks reveals functional architecture of rsfmri networks.
\newblock In \emph{ISMRM}, 2024{\natexlab{b}}.

\bibitem[Xie et~al.(2025{\natexlab{a}})Xie, Wang, Lin, Wen, Leong, and Wu]{intraandinter}
Linshan Xie, Xunda Wang, Xuehong Lin, Junjian Wen, Alex~TL Leong, and Ed~X Wu.
\newblock Intra-and inter-regional neural activity synchronizations drive rsfmri network dynamics upon single thalamic input.
\newblock In \emph{ISMRM}, 2025{\natexlab{a}}.

\bibitem[Xie et~al.(2025{\natexlab{b}})Xie, Wang, Lin, Wen, Ma, Leong, and Wu]{singlepulsenc}
Linshan Xie, Xunda Wang, Xuehong Lin, Junjian Wen, Teng Ma, Alex~TL Leong, and Ed~X Wu.
\newblock Brain-wide resting-state fmri network dynamics elicited by activation of single thalamic input.
\newblock \emph{Nature Communications}, 2025{\natexlab{b}}.

\bibitem[Xu et~al.(2025)Xu, Jin, Hao, Song, Sun, and Yuan]{llavacot}
Guowei Xu, Peng Jin, Li~Hao, Yibing Song, Lichao Sun, and Li~Yuan.
\newblock Llava-cot: Let vision language models reason step-by-step.
\newblock In \emph{ICCV}, 2025.

\bibitem[Xu et~al.(2024)Xu, Zhao, Zhou, Lin, Ng, and Feng]{pllava}
Lin Xu, Yilin Zhao, Daquan Zhou, Zhijie Lin, See~Kiong Ng, and Jiashi Feng.
\newblock Pllava: Parameter-free llava extension from images to videos for video dense captioning.
\newblock \emph{arXiv:2404.16994}, 2024.

\bibitem[Yan et~al.(2024)Yan, Li, He, Wang, Li, Li, Zeng, Wang, Wang, Qiao, et~al.]{videochattpo}
Ziang Yan, Zhilin Li, Yinan He, Chenting Wang, Kunchang Li, Xinhao Li, Xiangyu Zeng, Zilei Wang, Yali Wang, Yu~Qiao, et~al.
\newblock Task preference optimization: Improving multimodal large language models with vision task alignment.
\newblock \emph{arXiv:2412.19326}, 2024.

\bibitem[Yang et~al.(2022)Yang, Miech, Sivic, Laptev, and Schmid]{frozenbilm}
Antoine Yang, Antoine Miech, Josef Sivic, Ivan Laptev, and Cordelia Schmid.
\newblock Zero-shot video question answering via frozen bidirectional language models.
\newblock \emph{Advances in Neural Information Processing Systems}, 35:\penalty0 124--141, 2022.

\bibitem[Yang et~al.(2023)Yang, Li, Wang, Lin, Azarnasab, Ahmed, Liu, Liu, Zeng, and Wang]{mmreact}
Zhengyuan Yang, Linjie Li, Jianfeng Wang, Kevin Lin, Ehsan Azarnasab, Faisal Ahmed, Zicheng Liu, Ce~Liu, Michael Zeng, and Lijuan Wang.
\newblock Mm-react: Prompting chatgpt for multimodal reasoning and action.
\newblock \emph{arXiv:2303.11381}, 2023.

\bibitem[Yao et~al.(2023{\natexlab{a}})Yao, Yu, Zhao, Shafran, Griffiths, Cao, and Narasimhan]{tot}
Shunyu Yao, Dian Yu, Jeffrey Zhao, Izhak Shafran, Tom Griffiths, Yuan Cao, and Karthik Narasimhan.
\newblock Tree of thoughts: Deliberate problem solving with large language models.
\newblock In \emph{NeurIPS}, pp.\  11809--11822, 2023{\natexlab{a}}.

\bibitem[Yao et~al.(2023{\natexlab{b}})Yao, Zhao, Yu, Du, Shafran, Narasimhan, and Cao]{react}
Shunyu Yao, Jeffrey Zhao, Dian Yu, Nan Du, Izhak Shafran, Karthik Narasimhan, and Yuan Cao.
\newblock React: Synergizing reasoning and acting in language models.
\newblock In \emph{ICLR}, 2023{\natexlab{b}}.

\bibitem[Yu et~al.(2023)Yu, Cho, Yadav, and Bansal]{sevila}
Shoubin Yu, Jaemin Cho, Prateek Yadav, and Mohit Bansal.
\newblock Self-chained image-language model for video localization and question answering.
\newblock In \emph{NeurIPS}, pp.\  76749--76771, 2023.

\bibitem[Zhang et~al.(2023{\natexlab{a}})Zhang, Lu, Islam, Wang, Yu, Bansal, and Bertasius]{llovi}
Ce~Zhang, Taixi Lu, Md~Mohaiminul Islam, Ziyang Wang, Shoubin Yu, Mohit Bansal, and Gedas Bertasius.
\newblock A simple llm framework for long-range video question-answering.
\newblock \emph{arXiv:2312.17235}, 2023{\natexlab{a}}.

\bibitem[Zhang et~al.(2022)Zhang, Wu, and Li]{actionformer}
Chen-Lin Zhang, Jianxin Wu, and Yin Li.
\newblock Actionformer: Localizing moments of actions with transformers.
\newblock In \emph{ECCV}, pp.\  492--510, 2022.

\bibitem[Zhang et~al.(2023{\natexlab{b}})Zhang, Li, and Bing]{videollama}
Hang Zhang, Xin Li, and Lidong Bing.
\newblock Video-llama: An instruction-tuned audio-visual language model for video understanding.
\newblock \emph{arXiv:2306.02858}, 2023{\natexlab{b}}.

\bibitem[Zhang et~al.(2020{\natexlab{a}})Zhang, Sun, Jing, and Zhou]{vslnet}
Hao Zhang, Aixin Sun, Wei Jing, and Joey~Tianyi Zhou.
\newblock Span-based localizing network for natural language video localization.
\newblock \emph{arXiv:2004.13931}, 2020{\natexlab{a}}.

\bibitem[Zhang et~al.(2024)Zhang, Zhang, Li, Zeng, Yang, Zhang, Wang, Tan, Li, and Liu]{longva}
Peiyuan Zhang, Kaichen Zhang, Bo~Li, Guangtao Zeng, Jingkang Yang, Yuanhan Zhang, Ziyue Wang, Haoran Tan, Chunyuan Li, and Ziwei Liu.
\newblock Long context transfer from language to vision.
\newblock \emph{arXiv:2406.16852}, 2024.

\bibitem[Zhang et~al.(2020{\natexlab{b}})Zhang, Peng, Fu, and Luo]{2dtan}
Songyang Zhang, Houwen Peng, Jianlong Fu, and Jiebo Luo.
\newblock Learning 2d temporal adjacent networks for moment localization with natural language.
\newblock In \emph{AAAI}, pp.\  12870--12877, 2020{\natexlab{b}}.

\bibitem[Zhang et~al.(2023{\natexlab{c}})Zhang, Zhang, Li, Zhao, Karypis, and Smola]{mmcot}
Zhuosheng Zhang, Aston Zhang, Mu~Li, Hai Zhao, George Karypis, and Alex Smola.
\newblock Multimodal chain-of-thought reasoning in language models.
\newblock \emph{arXiv:2302.00923}, 2023{\natexlab{c}}.

\bibitem[Zhao et~al.(2023)Zhao, Misra, Krahenbuhl, and Girdhar]{lavila}
Yue Zhao, Ishan Misra, Philipp Krahenbuhl, and Rohit Girdhar.
\newblock Learning video representations from large language models.
\newblock In \emph{CVPR}, pp.\  6586--6597, 2023.

\bibitem[Zhou et~al.(2024)Zhou, Shu, Zhao, Wu, Xiao, Yang, Xiong, Zhang, Huang, and Liu]{mlvu}
Junjie Zhou, Yan Shu, Bo~Zhao, Boya Wu, Shitao Xiao, Xi~Yang, Yongping Xiong, Bo~Zhang, Tiejun Huang, and Zheng Liu.
\newblock Mlvu: A comprehensive benchmark for multi-task long video understanding.
\newblock \emph{arXiv:2406.04264}, 2024.

\bibitem[Zhu et~al.(2025)Zhu, Wang, Chen, Liu, Ye, Gu, Duan, Tian, Su, Shao, et~al.]{internvl3}
Jinguo Zhu, Weiyun Wang, Zhe Chen, Zhaoyang Liu, Shenglong Ye, Lixin Gu, Yuchen Duan, Hao Tian, Weijie Su, Jie Shao, et~al.
\newblock Internvl3: Exploring advanced training and test-time recipes for open-source multimodal models.
\newblock \emph{arXiv:2504.10479}, 2025.

\end{thebibliography}

\clearpage

\appendix

\section*{Appendix}

In this appendix, we provide more details about the model inference pipeline and implementation details to complement the main paper. Additional experiments, detailed analysis, and discussions are also incorporated. Below is the table of contents.

\begin{enumerate}[label=\textbf{\Alph*.}]
\item Model
\begin{enumerate}[label=\textbf{\arabic*.}]
\item Inference Pipeline
\item Implementation Details
\end{enumerate}
\item Experiments
\begin{enumerate}[label=\textbf{\arabic*.}]
\item Benchmarks and Settings
\item More Experimental Results
\item More Detailed Analysis
\end{enumerate}
\item Miscellaneous
\begin{enumerate}[label=\textbf{\arabic*.}]
\item Prompt Templates
\end{enumerate}
\item The Use of LLMs Statement
\end{enumerate}

\section{Model}

\subsection{Inference Pipeline}\label{sec:inference}

The formulation of VideoMind's inference pipeline is illustrated in Algorithm~\ref{algo:inerence}. Given a video $\mathcal{V}$ and a question $\mathcal{Q}$, the planner dynamically calls different roles on demand to analyze the multi-modal context and generate the answer.

\begin{center}
\vspace{-3mm}
\begin{minipage}{0.8\linewidth}
\begin{algorithm}[H]
\small
\caption{VideoMind's Chain-of-LoRA Pipeline}
\begin{algorithmic}[1]
\State \textbf{Input:} A video $\mathcal{V}$ and a question $\mathcal{Q}$
\State \textbf{Output:} An answer $\mathcal{A}$ to the question with temporal moment $\mathcal{T}=[t_s, t_e]$
\State Plan $\mathcal{P}$ $\leftarrow$ \texttt{\textbf{Planner}}$(\mathcal{V}, \mathcal{Q})$    
\If{\texttt{\textbf{Grounder}} $\in \mathcal{P}$}
\State $\{[t^i_s, t^i_e]\}_i$ $\leftarrow$ \texttt{\textbf{Grounder}}$(\mathcal{V}, \mathcal{Q})$
\ForAll{$i$}
\State $\tilde{\mathcal{V}_i}\leftarrow$ \texttt{ZoomIn}$(\mathcal{V}, [t^i_s, t^i_e])$
\State $Score_i\leftarrow$ \texttt{\textbf{Verifier}}$(\tilde{\mathcal{V}_i}, \mathcal{Q})$
\EndFor
\State $i\leftarrow \arg \max s_i (Score_i)$
\EndIf        
\If{\texttt{\textbf{Answerer}} $\in \mathcal{P}$}
\State $\mathcal{A}\leftarrow$\texttt{\textbf{Answerer}}$(\tilde{\mathcal{V}_i}, \mathcal{Q})$
\EndIf
\State \Return $(\mathcal{A}, \mathcal{T})$
\end{algorithmic}
\label{algo:inerence}
\end{algorithm}
\end{minipage}
\vspace{2mm}
\end{center}

\subsection{Implementation Details}\label{sec:implementation}

We leverage the 2B and 7B versions of Qwen2-VL \citep{qwen2vl} as our base models, and apply LoRA adapters with $\mathrm{rank} = 64$ and $\mathrm{alpha} = 64$ to the planner, grounder, and verifier. The hidden size of the timestamp decoder is set to 256. The maximum number of tokens per frame and maximum number of frames for the planner, grounder, verifier, and answerer are set as [64, 100], [64, 150], [64, 64], and [256, 32], respectively. We train different roles separately on different datasets and load them together during inference, so that the model can efficiently switch roles by activating different LoRAs. During training, we set the global batch size to 32, and utilize the AdamW optimizer \citep{adamw} with learning rates of 2e-5, 1e-4, and 5e-5 for planner, grounder, and verifier, respectively. All the roles were trained for 1 epoch on their specific datasets, with a linear warmup in the first 3\% steps. During inference, we apply an NMS with $\mathrm{IoU} = 0.75$ to reduce duplicated moments from the grounder.

\section{Experiments}

\subsection{Benchmarks and Settings}\label{sec:benchmarks}

The experiments are extensively designed across 15 diverse benchmarks. The statistics are listed in Table~\ref{tab:benchmark}. The major benchmarks are introduced below.

\textbf{CG-Bench \citep{cgbench}} is designed for long video grounded question answering, featuring a diverse domain and various evaluation metrics. It includes 1.2K manually curated videos, ranging from 10 to 80 minutes, with a total of 12K QA pairs. The dataset is categorized into perception, reasoning, and hallucination question types, and introduces clue-based evaluation methods like white box and black box assessments to ensure models provide answers based on accurate video reasoning.

\begin{table}
\centering
\scriptsize
\setlength{\tabcolsep}{7.65pt}
\caption{Details of the evaluation benchmarks. The datasets encompass three representative tasks, \ie, Grounded VideoQA, Video Temporal Grounding, and General VideoQA, with video durations ranging from several seconds to more than one hour.}
\vspace{2mm}
\begin{tabularx}{0.845\linewidth}{l|r|c|c}
\toprule
\textbf{Dataset} & \textbf{Duration} & \textbf{Domain} & \textbf{Main Metrics} \\
\midrule
\multicolumn{4}{l}{\textit{\textcolor{gray}{Grounded Video Question Answering (Grounding + QA)}}} \\
\midrule
CG-Bench \citep{cgbench} & 1624.4s & Diverse & rec.@IoU, acc.@IoU \\
ReXTime \citep{rextime} & 141.1s & Vlog, News, Activity & mIoU, Acc\,(IoU\,$\geqslant$\,0.5) \\
NExT-GQA \citep{nextgqa} & 39.5s & Reasoning & mIoP, Acc@GQA \\
\midrule
\multicolumn{4}{l}{\textit{\textcolor{gray}{Video Temporal Grounding (Grounding only)}}} \\
\midrule 
Charades-STA \citep{charadessta} & 30.1s & Indoor & R@\{0.3\,$\sim$\,0.7\}, mIoU \\
ActivityNet-Captions \citep{activitynetcaptions} & 111.4s & Activity & R@\{0.3\,$\sim$\,0.7\}, mIoU \\
QVHighlights \citep{qvhighlights} & 150s & Vlog, News & R@\{0.5,\,0.7\}, mAP \\
TACoS \citep{tacos} & 358.2s & Cooking & R@\{0.3\,$\sim$\,0.7\}, mIoU \\
Ego4D-NLQ \citep{ego4d} & 379.0s & Egocentric & R@\{0.3\,$\sim$\,0.7\}, mIoU \\
ActivityNet-RTL \citep{lita} & 111.4s & Reasoning & P@0.5, mIoU \\
\midrule
\multicolumn{4}{l}{\textit{\textcolor{gray}{General Video Question Answering (QA only)}}} \\
\midrule
Video-MME \citep{videomme} & 1017.9s & Diverse & Acc (\textit{w/o subs}) \\
MLVU \citep{mlvu} & 930s & Diverse & Acc \\
LVBench \citep{lvbench} & 4101s & Diverse & Acc \\
MVBench \citep{mvbench} & 15s & Diverse & Acc \\
LongVideoBench \citep{longvideobench} & 473s & Diverse & Acc \\
\bottomrule
\end{tabularx}
\label{tab:benchmark}
\vspace{-4mm}
\end{table}

\begin{table}
\centering
\scriptsize
\setlength{\tabcolsep}{7.2pt}
\caption{Performance on MultiHop-EgoQA \citep{multihopegoqa}. \underline{FT} means fine-tuning on the target dataset. \underline{Sent. Sim.} denotes sentence similarity computed by \texttt{all-MiniLM-L6-v2}.}
\vspace{2mm}
\begin{tabularx}{0.83\linewidth}{l|cc|cc|cc}
\toprule
\multirow{2.6}{*}{\textbf{Method}} & \multirow{2.6}{*}{\textbf{Size}} & \multirow{2.6}{*}{\textbf{FT}} & \multicolumn{2}{c|}{\textbf{Temporal Grounding}} & \multicolumn{2}{c}{\textbf{Question Answering}} \\
\cmidrule{4-5} \cmidrule{6-7}
&&& IoU@0.3 & mIoU & Sent. Sim. & Score \\
\midrule
\textcolor{gray}{Human} & -- & -- & \textcolor{gray}{87.0} & \textcolor{gray}{61.8} & \textcolor{gray}{74.3} & \textcolor{gray}{7.5} \\
\midrule
GPT-4o \citep{gpt4o} & -- & \xmark & 12.0 & 12.2 & 73.7 & \textbf{5.4} \\
InternVL2 \citep{internvl2} & 8B & \xmark & 6.3 & 6.6 & 71.9 & 4.5 \\
LLaVA-NeXT-Video \citep{llavanext} & 7B & \xmark & -- & -- & 62.1 & 4.2 \\
TimeChat \citep{timechat} & 7B & \xmark & 3.0 & 3.6 & 58.9 & 3.3 \\
VTimeLLM \citep{vtimellm} & 7B & \xmark & 8.8 & 9.2 & 70.5 & 4.3 \\
GeLM \citep{multihopegoqa} & 7B & \cmark & 18.2 & 16.7 & \underline{75.0} & 4.8 \\
\midrule
\rowcolor{blue!7.5} \textbf{VideoMind} (Ours) & 2B & \xmark & \underline{23.2} & \underline{17.8} & 58.8 & 3.5 \\
\rowcolor{blue!7.5} \textbf{VideoMind} (Ours) & 7B & \xmark & \textbf{25.1} & \textbf{19.0} & \textbf{77.3} & \underline{4.9} \\
\bottomrule
\end{tabularx}
\label{tab:multihopegoqa}
\end{table}

\textbf{ReXTime \citep{rextime}} tests models on complex temporal reasoning, using an automated pipeline for QA pair generation, significantly reducing manual effort. It includes 921 validation and 2.1K test samples, each manually curated for accuracy, and highlights a 14.3\% accuracy gap between SoTA models and human performance. This benchmark is crucial for evaluating models on cause-and-effect relationships across video segments.

\textbf{NExT-GQA \citep{nextgqa}} aims to challenge models to reason about causal and temporal actions, supporting both multi-choice and open-ended tasks. This is an extension of NExT-QA \citep{nextqa} comprising 10.5K manually labeled video QA pairs with temporal segments. The samples in this benchmark are from ``causal'' and ``temporal'' classes, while the ``descriptive'' questions in NExT-QA are discarded.

\noindent\textbf{Charades-STA \citep{charadessta}} contains 10K in-door videos, averaging 30.1 seconds each, with 16K temporal annotations spanning daily activity, alongside free-text descriptions. These rich annotations make Charades-STA particularly suitable for evaluating temporal grounding models under indoor environments.

\textbf{ActivityNet-Captions \citep{activitynetcaptions}} is a large-scale benchmark with 20K untrimmed YouTube videos with a total of 849 hours, covering diverse activities from personal care to sports. This dataset contains high-quality dense video captioning annotations (3.65 temporally localized sentences per video), which we use as queries for video temporal grounding. Each query has an average length of 13.5 words.

\begin{table}
\centering
\scriptsize
\setlength{\tabcolsep}{5pt}
\caption{Video temporal grounding on TACoS \citep{tacos}. \underline{FT} means fine-tuning on the target dataset. Note that our method was co-trained on this dataset.}
\vspace{2mm}
\begin{tabularx}{0.6615\linewidth}{l|cc|cccc}
\toprule
\textbf{Method} & \textbf{Size} & \textbf{FT} & \textbf{R@0.3} & \textbf{R@0.5} & \textbf{R@0.7} & \textbf{mIoU} \\
\midrule
\multicolumn{6}{l}{\textit{\textcolor{gray}{Non-LLM-based Specialists}}} \\
\midrule
2D-TAN \citep{2dtan} & -- & \cmark & 40.0 & 28.0 & 12.9 & 27.2 \\
Moment-DETR \citep{qvhighlights} & -- & \cmark & 38.0 & 24.7 & 12.0 & 25.5 \\
UniVTG \citep{univtg} & -- & \cmark & 51.4 & 35.0 & 17.4 & 33.6 \\
R$^2$-Tuning \citep{r2tuning} & -- & \cmark & 49.7 & 38.7 & 25.1 & 35.9 \\
\midrule
\multicolumn{6}{l}{\textit{\textcolor{gray}{LLM-based Generalists}}} \\
\midrule
\rowcolor{blue!7.5} \textbf{VideoMind} (Ours) & 2B & \xmark & \underline{38.6} & \underline{26.9} & \underline{15.5} & \underline{27.4} \\
\rowcolor{blue!7.5} \textbf{VideoMind} (Ours) & 7B & \xmark & \textbf{49.5} & \textbf{36.2} & \textbf{21.4} & \textbf{34.4} \\
\bottomrule
\end{tabularx}
\label{tab:tacos}
\vspace{-2mm}
\end{table}

\begin{table}[t!]
\centering
\scriptsize
\setlength{\tabcolsep}{5pt}
\caption{Performance of video temporal grounding on Ego4D-NLQ \citep{ego4d}. \underline{FT} means fine-tuning on the target dataset. VideoMind-Ego is a variant of our method trained with extra 67K egocentric grounding samples from NaQ \citep{ego4dnaq}.}
\vspace{2mm}
\begin{tabularx}{0.6615\linewidth}{l|cc|cccc}
\toprule
\textbf{Method} & \textbf{Size} & \textbf{FT} & \textbf{R@0.3} & \textbf{R@0.5} & \textbf{R@0.7} & \textbf{mIoU} \\
\midrule
\multicolumn{6}{l}{\textit{\textcolor{gray}{Non-LLM-based Specialists}}} \\
\midrule
2D-TAN \citep{2dtan} & -- & \cmark & 4.3 & 1.8 & 0.6 & 3.4 \\
VSLNet \citep{vslnet} & -- & \cmark & 4.5 & 2.4 & 1.0 & 3.5 \\
Moment-DETR \citep{qvhighlights} & -- & \cmark & 4.3 & 1.8 & 0.7 & 3.5 \\
UniVTG \citep{univtg} & -- & \cmark & 7.3 & 4.0 & 1.3 & 4.9 \\
R$^2$-Tuning \citep{r2tuning} & -- & \cmark & 7.2 & 4.5 & 2.1 & 4.9 \\
UniVTG \citep{univtg} & -- & \xmark & 6.5 & 3.5 & 1.2 & 4.6 \\
\midrule
\multicolumn{6}{l}{\textit{\textcolor{gray}{LLM-based Generalists}}} \\
\midrule
\rowcolor{blue!7.5} \textbf{VideoMind} (Ours) & 2B & \xmark & \underline{5.9} & 2.9 & 1.2 & 4.7 \\
\rowcolor{blue!7.5} \textbf{VideoMind} (Ours) & 7B & \xmark & \textbf{7.2} & \underline{3.7} & \underline{1.7} & \textbf{5.4} \\
\midrule
\rowcolor{blue!7.5} \textbf{VideoMind-Ego} (Ours) & 2B & \xmark & \textbf{7.2} & \textbf{3.9} & \textbf{1.8} & \underline{5.3} \\
\bottomrule
\end{tabularx}
\label{tab:ego4d}
\vspace{-2mm}
\end{table}

\subsection{More Experimental Results}

\paragraph{Multi-Hop Grounded Question Answering} To investigate the performance of our method on novel tasks that require a hybrid or dynamically generated sequence of steps, we evaluate our method on MultiHop-EgoQA \citep{multihopegoqa}, a Grounded VideoQA dataset highlighting multi-hop temporal reasoning. For each question, the model must ground and reason on multiple relevant moments before answering, which is a paradigm that does not fit neatly into the pre-defined single-hop grounding pipeline. The evaluation results are shown in Table~\ref{tab:multihopegoqa}. Thanks to VideoMind's architectural design to produce multiple candidate moments in a single grounding step, it can effectively capture the multi-hop evidence required by this benchmark. As a result, our method achieves strong zero-shot performance, surpassing all open-source baselines and remaining competitive to closed-source GPT-4o \citep{gpt4o} across both grounding metrics and QA metrics.

\begin{table}
\centering
\scriptsize
\begin{minipage}{0.58\textwidth}
\setlength{\tabcolsep}{3.025pt}
\caption{Fine-tuned video temporal grounding results on QVHighlights \citep{qvhighlights}.}
\vspace{2mm}
\begin{tabularx}{\linewidth}{l|c|cc|ccc}
\toprule
\multirow{2.6}{*}{\textbf{Method}} & \multirow{2.6}{*}{\textbf{Size}} & \multicolumn{2}{c|}{\textbf{R1}} & \multicolumn{3}{c}{\textbf{mAP}} \\
\cmidrule{3-4} \cmidrule{5-7}
&& @0.5 & @0.7 & @0.5 & @0.75 & Avg. \\
\midrule
\multicolumn{7}{l}{\textit{\textcolor{gray}{Non-LLM-based Specialists}}} \\
\midrule
XML \citep{tvr} & -- & 41.83 & 30.35 & 44.63 & 31.73 & 32.14 \\
XML+ \citep{qvhighlights} & -- & 46.69 & 33.46 & 47.89 & 34.67 & 34.90 \\
Moment-DETR \citep{qvhighlights} & -- & 59.78 & 40.33 & 60.51 & 35.36 & 36.14 \\
UMT \citep{umt} & -- & 60.83 & 43.26 & 57.33 & 39.12 & 38.08 \\
MomentDiff \citep{momentdiff} & -- & 58.21 & 41.48 & 54.57 & 37.21 & 36.84 \\
QD-DETR \citep{qddetr} & -- & 62.40 & 44.98 & 62.52 & 39.88 & 39.86 \\
UniVTG \citep{univtg} & -- & 65.43 & 50.06 & 64.06 & 45.02 & 43.63 \\
R$^2$-Tuning \citep{r2tuning} & -- & 68.03 & 49.35 & 69.04 & 47.56 & 46.17 \\
\midrule
\multicolumn{7}{l}{\textit{\textcolor{gray}{LLM-based Generalists}}} \\
\midrule
\rowcolor{blue!7.5} \textbf{VideoMind} (Ours) & 2B & \underline{75.42} & \underline{59.35} & \underline{74.11} & \underline{55.15} & \underline{51.60} \\
\rowcolor{blue!7.5} \textbf{VideoMind} (Ours) & 7B & \textbf{78.53} & \textbf{61.09} & \textbf{76.07} & \textbf{58.17} & \textbf{54.19} \\
\bottomrule
\end{tabularx}
\label{tab:qvhighlights}
\end{minipage}
\hfill
\begin{minipage}{0.4\textwidth}
\vspace{11.4mm}
\setlength{\tabcolsep}{2.35pt}
\caption{Comparison of performance on reasoning temporal localization on ActivityNet-RTL \citep{lita}. Our zero-shot VideoMind-7B outperforms the strong fine-tuned baseline LITA-13B \citep{lita} by a considerable margin.}
\vspace{2mm}
\begin{tabularx}{\linewidth}{l|cc|cc}
\toprule
\textbf{Method} & \textbf{Size} & \textbf{FT} & \textbf{P@0.5} & \textbf{mIoU} \\
\midrule
LITA \citep{lita} & 7B & \cmark & 21.2 & 24.1 \\
LITA \citep{lita} & 13B & \cmark & 25.9 & 28.6 \\
\midrule
\rowcolor{blue!7.5} \textbf{VideoMind} (Ours) & 2B & \xmark & \underline{20.1} & \underline{22.7} \\
\rowcolor{blue!7.5} \textbf{VideoMind} (Ours) & 7B & \xmark & \textbf{28.0} & \textbf{31.3} \\
\bottomrule
\end{tabularx}
\label{tab:rtl}
\end{minipage}
\vspace{-3mm}
\end{table}

\begin{table}
\centering
\scriptsize
\setlength{\tabcolsep}{5pt}
\caption{Performance of VideoQA on LongVideoBench \citep{longvideobench} \texttt{val} split.}
\vspace{2mm}
\begin{tabularx}{0.765\linewidth}{l|c|c|cccc}
\toprule
\multirow{2.6}{*}{\textbf{Method}} & \multirow{2.6}{*}{\textbf{Size}} & \multirow{2.6}{*}{\textbf{Acc}} & \multicolumn{4}{c}{\textbf{Acc @ Duration Groups}} \\
\cmidrule{4-7}
&&& (8,\,15] & (15,\,60] & (180,\,600] & (900,\,3600] \\
\midrule
\textcolor{gray}{GPT-4o \citep{gpt4o}} & \textcolor{gray}{--} & \textcolor{gray}{66.7} & \textcolor{gray}{71.4} & \textcolor{gray}{76.7} & \textcolor{gray}{69.1} & \textcolor{gray}{60.9} \\
\textcolor{gray}{GPT-4 Turbo \citep{gpt4}} & \textcolor{gray}{--} & \textcolor{gray}{59.0} & \textcolor{gray}{65.2} & \textcolor{gray}{68.2} & \textcolor{gray}{62.4} & \textcolor{gray}{50.5} \\
\textcolor{gray}{Gemini-1.5-Pro \citep{gemini1.5}} & \textcolor{gray}{--} & \textcolor{gray}{64.0} & \textcolor{gray}{67.4} & \textcolor{gray}{75.1} & \textcolor{gray}{65.3} & \textcolor{gray}{58.6} \\
\textcolor{gray}{Gemini-1.5-Flash \citep{gemini1.5}} & \textcolor{gray}{--} & \textcolor{gray}{61.6} & \textcolor{gray}{68.3} & \textcolor{gray}{76.2} & \textcolor{gray}{62.6} & \textcolor{gray}{54.0} \\
\midrule
Idefics2 \citep{idefics2} & 8B & \underline{49.7} & \underline{59.8} & \underline{65.7} & 47.8 & \underline{42.7} \\
Phi-3-Vision \citep{phi3} & 4B & 49.6 & 59.3 & 61.6 & 46.8 & 44.7 \\
Mantis-Idefics2 \citep{mantis} & 8B & 47.0 & 56.6 & 55.8 & 45.6 & 42.2 \\
Mantis-BakLLaVA \citep{mantis} & 7B & 43.7 & 53.4 & 57.6 & 40.3 & 38.7 \\
\midrule
\rowcolor{blue!7.5} \textbf{VideoMind} (Ours) & 2B & 48.8 & 59.3 & 59.3 & \underline{49.3} & 41.7 \\
\rowcolor{blue!7.5} \textbf{VideoMind} (Ours) & 7B & \textbf{56.3} & \textbf{67.7} & \textbf{67.4} & \textbf{56.8} & \textbf{48.6} \\
\bottomrule
\end{tabularx}
\label{tab:longvideobench}
\vspace{-3mm}
\end{table}

\begin{table}[h!]
\centering
\fontsize{5.8pt}{6.8pt}\selectfont
\setlength{\tabcolsep}{1.8pt}
\caption{Performance comparison on general VideoQA on MVBench \citep{mvbench}.}
\vspace{2mm}
\begin{tabularx}{\linewidth}{l|c|cccccccccccccccccccc|c}
\toprule
\textbf{Model} & \textbf{Size} & \textbf{AS} & \textbf{AP} & \textbf{AA} & \textbf{FA} & \textbf{UA} & \textbf{OE} & \textbf{OI} & \textbf{OS} & \textbf{MD} & \textbf{AL} & \textbf{ST} & \textbf{AC} & \textbf{MC} & \textbf{MA} & \textbf{SC} & \textbf{FP} & \textbf{CO} & \textbf{EN} & \textbf{ER} & \textbf{CI} & \textbf{Avg.} \\
\midrule
GPT-4V \citep{gpt4v} & -- & 55.5 & 63.5 & 72.0 & 46.5 & \underline{73.5} & 18.5 & 59.0 & 29.5 & 12.0 & 40.5 & 83.5 & 39.0 & 12.0 & 22.5 & 45.0 & 47.5 & 52.0 & 31.0 & \textbf{59.0} & 11.0 & 43.5 \\
\midrule
Video-ChatGPT \citep{videochatgpt} & 7B & 23.5 & 26.0 & 62.0 & 22.5 & 26.5 & 54.0 & 28.0 & 40.0 & 23.0 & 20.0 & 31.0 & 30.5 & 25.5 & 39.5 & 48.5 & 29.0 & 33.0 & 29.5 & 26.0 & 35.5 & 32.7 \\
Video-LLaMA \citep{videollama} & 7B & 27.5 & 25.5 & 51.0 & 29.0 & 39.0 & 48.0 & 40.5 & 38.0 & 22.5 & 22.5 & 43.0 & 34.0 & 22.5 & 32.5 & 45.5 & 32.5 & 40.0 & 30.0 & 21.0 & 37.0 & 34.1 \\
VideoChat \citep{videochat} & 7B & 33.5 & 26.5 & 56.0 & 33.5 & 40.5 & 53.0 & 40.5 & 30.0 & 25.5 & 27.0 & 48.5 & 35.0 & 20.5 & 42.5 & 46.0 & 26.5 & 41.0 & 23.5 & 23.5 & 36.0 & 35.5 \\
Video-LLaVA \citep{videollava} & 7B & 46.0 & 42.5 & 56.5 & 39.0 & 53.5 & 53.0 & 48.0 & \underline{41.0} & 29.0 & 31.5 & 82.5 & \underline{45.0} & 26.0 & 53.0 & 41.5 & 33.5 & 41.5 & 27.5 & 38.5 & 31.5 & 43.0 \\
TimeChat \citep{timechat} & 7B & 40.5 & 36.0 & 61.0 & 32.5 & 53.0 & 53.5 & 41.5 & 29.0 & 19.5 & 26.5 & 66.5 & 34.0 & 20.0 & 43.5 & 42.0 & 36.5 & 36.0 & 29.0 & 35.0 & 35.0 & 38.5 \\
PLLaVA \citep{pllava} & 7B & 58.0 & 49.0 & 55.5 & 41.0 & 61.0 & 56.0 & 61.0 & 36.0 & 23.5 & 26.0 & 82.0 & 39.5 & 42.0 & 52.0 & 45.0 & 42.0 & 53.5 & 30.5 & 48.0 & 31.0 & 46.6 \\
ShareGPT4Video \citep{sharegpt4video} & 7B & 49.5 & 39.5 & 79.5 & 40.0 & 54.5 & 82.5 & 54.5 & 32.5 & 50.5 & 41.5 & 84.5 & 35.5 & 62.5 & 75.0 & \textbf{51.0} & 25.5 & 46.5 & 28.5 & 39.0 & 51.5 & 51.2 \\
ST-LLM \citep{stllm} & 7B & 66.0 & 53.5 & \textbf{84.0} & 44.0 & 58.5 & 80.5 & 73.5 & 38.5 & 42.5 & 31.0 & 86.5 & 36.5 & 56.5 & 78.5 & 43.0 & 44.5 & 46.5 & 34.5 & 41.5 & 58.5 & 54.9 \\
VideoGPT+ \citep{videogpt} & 3.8B & 69.0 & 60.0 & 83.0 & 48.5 & 66.5 & 85.5 & 75.5 & 36.0 & 44.0 & 34.0 & 89.5 & 39.5 & 71.0 & 90.5 & 45.0 & 53.0 & 50.0 & 29.5 & 44.0 & 60.0 & 58.7 \\
VideoChat2 \citep{mvbench} & 7B & \underline{75.5} & 58.0 & \underline{83.5} & \textbf{50.5} & 60.5 & 87.5 & \underline{74.5} & \textbf{45.0} & 47.5 & \underline{44.0} & 82.5 & 37.0 & 64.5 & 87.5 & \textbf{51.0} & \textbf{66.5} & 47.0 & 35.0 & 37.0 & \textbf{72.5} & 60.4 \\
\midrule
\rowcolor{blue!7.5} \textbf{VideoMind} (Ours) & 2B & \textbf{78.5} & \textbf{76.0} & 75.5 & 46.0 & 69.5 & \underline{90.5} & 71.5 & 33.0 & \underline{48.0} & 40.0 & \textbf{92.5} & \textbf{52.5} & \underline{71.5} & \underline{92.0} & 44.5 & \underline{61.5} & \underline{61.5} & \underline{37.5} & 51.0 & 57.0 & \underline{62.5} \\
\rowcolor{blue!7.5} \textbf{VideoMind} (Ours) & 7B & 74.0 & \underline{71.5} & 81.0 & \underline{50.0} & \textbf{77.0} & \textbf{93.0} & \textbf{75.0} & 38.0 & \textbf{48.5} & \textbf{46.0} & \underline{91.0} & 39.0 & \textbf{80.0} & \textbf{94.5} & \underline{49.5} & 55.5 & \textbf{70.0} & \textbf{40.5} & \underline{57.0} & \underline{61.0} & \textbf{64.6} \\
\bottomrule
\end{tabularx}
\label{tab:mvbench}
\vspace{-3mm}
\end{table}

\paragraph{Video Temporal Grounding} We additionally compare VideoMind with representative methods on the challenging TACoS \citep{tacos}, Ego4D-NLQ \citep{ego4d}, and QVHighlights \citep{qvhighlights} datasets in Table~\ref{tab:tacos}, Table~\ref{tab:ego4d}, and Table~\ref{tab:qvhighlights}, respectively. Our 2B model performs better than the strong task-specific baseline UniVTG \citep{univtg} on TACoS but slightly worse than it on Ego4D-NLQ. This is justifiable as neither the grounder nor the verifier was trained on egocentric videos, while UniVTG was pretrained on 1.8M samples from Ego4D \citep{ego4d}. To align the setting, we trained an additional VideoMind-2B variant with extra 67K grounding samples from NaQ \citep{ego4dnaq}. To our best knowledge, VideoMind is \textbf{the first LLM-based grounding model that supports multi-moment outputs}, thereby being able to be evaluated on QVHighlights. Compared with task-specific experts, our VideoMind-2B significantly outperforms all previous methods and sets a new state-of-the-art.

\paragraph{Reasoning Temporal Localization} We also evaluate the generalizability of grounder and verifier on the more challenging reasoning temporal localization \citep{lita} task, which is similar to video temporal grounding, but the queries are not directly describing the moment. The models are required to infer the actual event using their world knowledge. The results in Table~\ref{tab:rtl} show that VideoMind can successfully generalize its zero-shot grounding capability to complex scenarios.

\paragraph{General Video Question Answering} For the task of long VideoQA, we also provide evaluations on LongVideoBench \citep{longvideobench} in Table~\ref{tab:longvideobench}, which further verifies the effectiveness of VideoMind on videos scaling to one-hour long. Table~\ref{tab:mvbench} presents more results of VideoMind on MVBench \citep{mvbench}, which is a benchmark with very short videos (around 15s). Our model can still achieve good performance on these short video scenarios.

\paragraph{Comparison with Text-based Reasoning Models} In Table~\ref{tab:reasoning}, we compare our method with representative pure text-based video reasoning methods. Our method significantly outperforms both baselines on all benchmarks, demonstrating that vision-centric reasoning is superior to pure text-based reasoning on long/complex video reasoning tasks.

\begin{table}
\centering
\scriptsize
\setlength{\tabcolsep}{4pt}
\caption{Comparison with representative video reasoning methods on video QA/grounding tasks.}
\vspace{2mm}
\begin{tabularx}{0.9\linewidth}{l|c|c|c|c|cc|cc}
\toprule
\multirow{2.6}{*}{\textbf{Method}} & \multirow{2.6}{*}{\textbf{Size}} & \textbf{CG-Bench} & \textbf{MLVU} & \textbf{LVBench} & \multicolumn{2}{c|}{\textbf{Charades-STA}} & \multicolumn{2}{c}{\textbf{ActivityNet-Captions}} \\
\cmidrule{3-9}
&& long-acc. & M-Avg & Overall & R@0.5 & mIoU & R@0.5 & mIoU \\
\midrule
\multicolumn{9}{l}{\textit{\textcolor{gray}{Pure Text-based Reasoning Models}}} \\
\midrule
LongVILA-R1 \citep{longvilar1} & 7B & 26.7 & 56.5 & 34.7 & 30.3 & 30.0 & 16.4 & 21.4 \\
Video-R1 \citep{videor1} & 7B & 34.4 & 63.1 & 38.4 & 35.3 & 34.9 & 22.6 & 28.0 \\
\midrule
\multicolumn{9}{l}{\textit{\textcolor{gray}{Vision-centric Reasoning Models}}} \\
\midrule
\rowcolor{blue!7.5} \textbf{VideoMind} (Ours) & 7B & \textbf{38.4} & \textbf{64.4} & \textbf{40.8} & \textbf{59.1} & \textbf{50.2} & \textbf{30.3} & \textbf{33.3} \\
\bottomrule
\end{tabularx}
\label{tab:reasoning}
\vspace{-3mm}
\end{table}

\begin{table}[t!]
\centering
\scriptsize
\setlength{\tabcolsep}{6.65pt}
\caption{Performance of different timestamp modeling designs on Charades-STA \citep{charadessta}.}
\vspace{2mm}
\begin{tabularx}{0.65\linewidth}{l|ccccc}
\toprule
\textbf{Method} & \textbf{R@0.3} & \textbf{R@0.5} & \textbf{R@0.7} & \textbf{mIoU} \\
\midrule
Text-only \citep{timechat} & 56.8 & 39.5 & 14.3 & 36.1 \\
Special Tokens \citep{momentor} & 56.4 & 39.2 & 14.5 & 35.7 \\
Embedding Matching \citep{etbench} & 59.6 & 43.5 & 17.0 & 38.2 \\
Time Marker \citep{timemarker} & \underline{60.5} & \underline{43.9} & \underline{17.2} & \underline{38.6} \\
\rowcolor{blue!7.5} \textbf{Timestamp Decoder} (Ours) & \textbf{64.1} & \textbf{47.2} & \textbf{21.7} & \textbf{42.0} \\
\bottomrule
\end{tabularx}
\label{tab:grounder}
\vspace{-3mm}
\end{table}

\begin{table}[t!]
\centering
\scriptsize
\setlength{\tabcolsep}{2.95pt}
\caption{Case distribution on ReXTime \citep{rextime} and NExT-GQA \citep{nextgqa}. \textit{Correct}, \textit{Planning}, \textit{Grounding}, \textit{Verification}, and \textit{Answering} refers to correct prediction, planning error, grounding error, verification error, and answering error, respectively.}
\vspace{2mm}
\begin{tabularx}{\linewidth}{l|c|ccccc|ccccc}
\toprule
\multirow{2.6}{*}{\textbf{Method}} & \multirow{2.6}{*}{\textbf{Size}} & \multicolumn{5}{c|}{\textbf{ReXTime}} & \multicolumn{5}{c}{\textbf{NExT-GQA}} \\
\cmidrule{3-12}
&& Correct & Planning & Grounding & Verification & Answering & Correct & Planning & Grounding & Verification & Answering \\
\midrule
\textbf{VideoMind} & 2B & 69.1\% & 1.2\% & 18.3\% & 5.7\% & 5.7\% & 71.2\% & 1.9\% & 14.0\% & 6.9\% & 6.0\% \\
\textbf{VideoMind} & 7B & 74.6\% & 1.1\% & 15.0\% & 4.6\% & 4.7\% & 76.6\% & 0.7\% & 11.8\% & 5.8\% & 5.1\% \\
\bottomrule
\end{tabularx}
\label{tab:case}
\vspace{2mm}
\end{table}

\begin{figure}[t!]
\centering
\begin{minipage}[b]{0.22\textwidth}
\centering
\includegraphics[width=\linewidth]{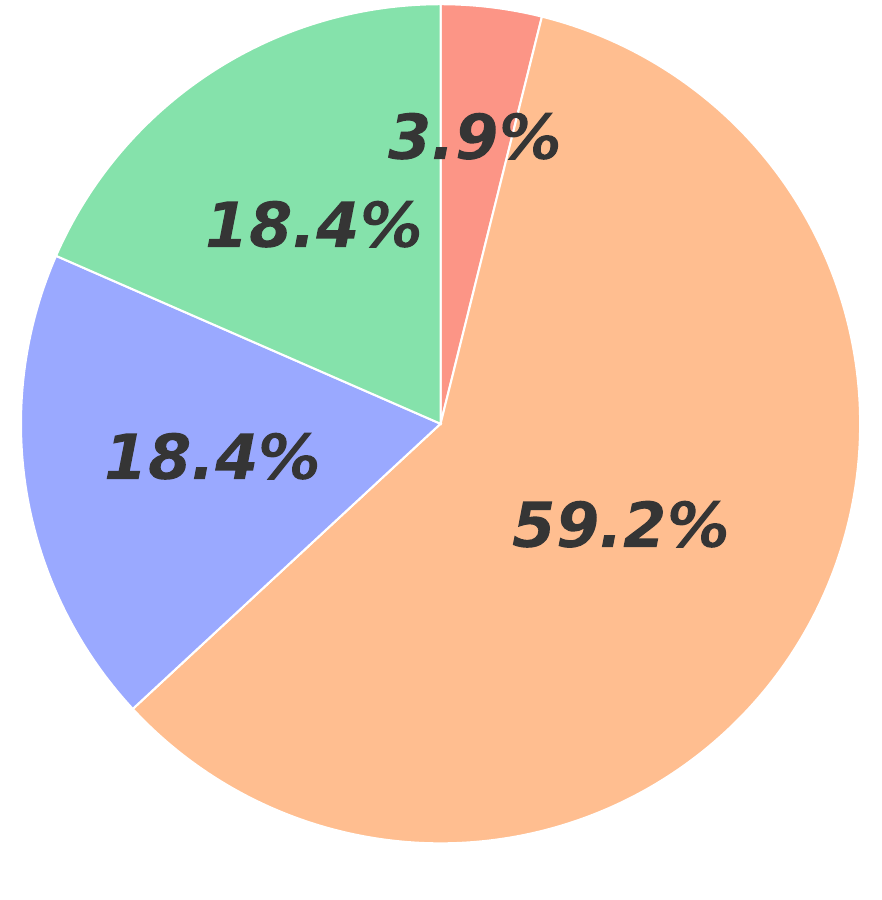}
(a) ReXTime (2B)
\end{minipage}
\hfill
\begin{minipage}[b]{0.22\textwidth}
\centering
\includegraphics[width=\linewidth]{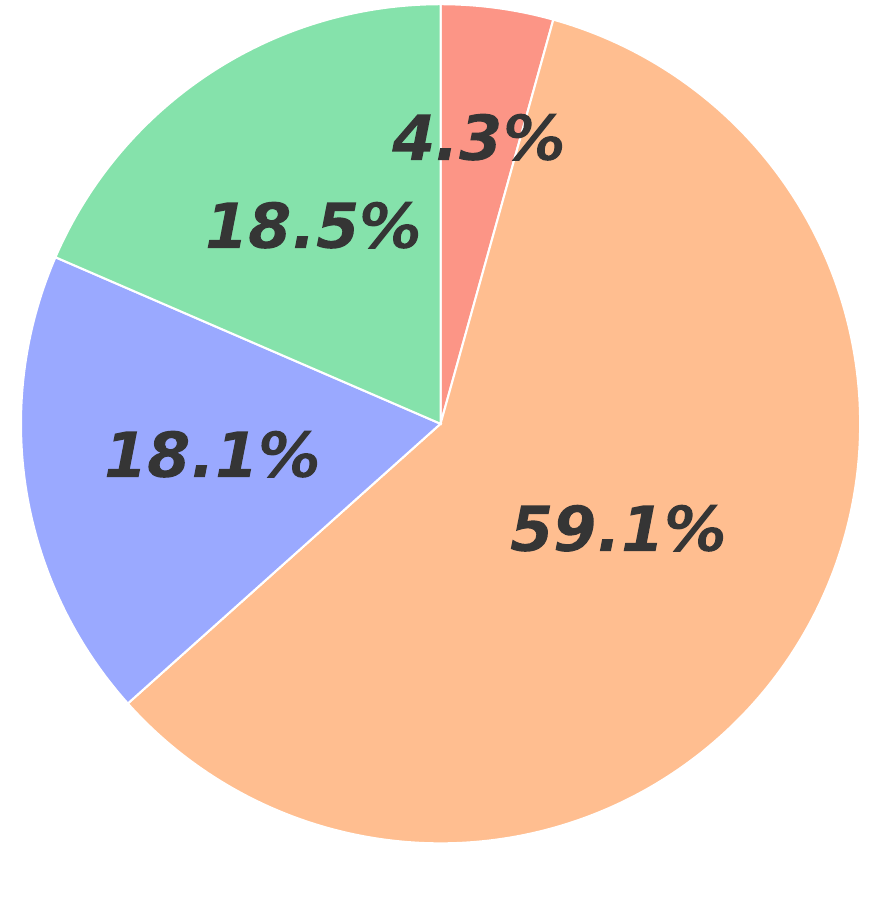}
(b) ReXTime (7B)
\end{minipage}
\hfill
\begin{minipage}[b]{0.22\textwidth}
\centering
\includegraphics[width=\linewidth]{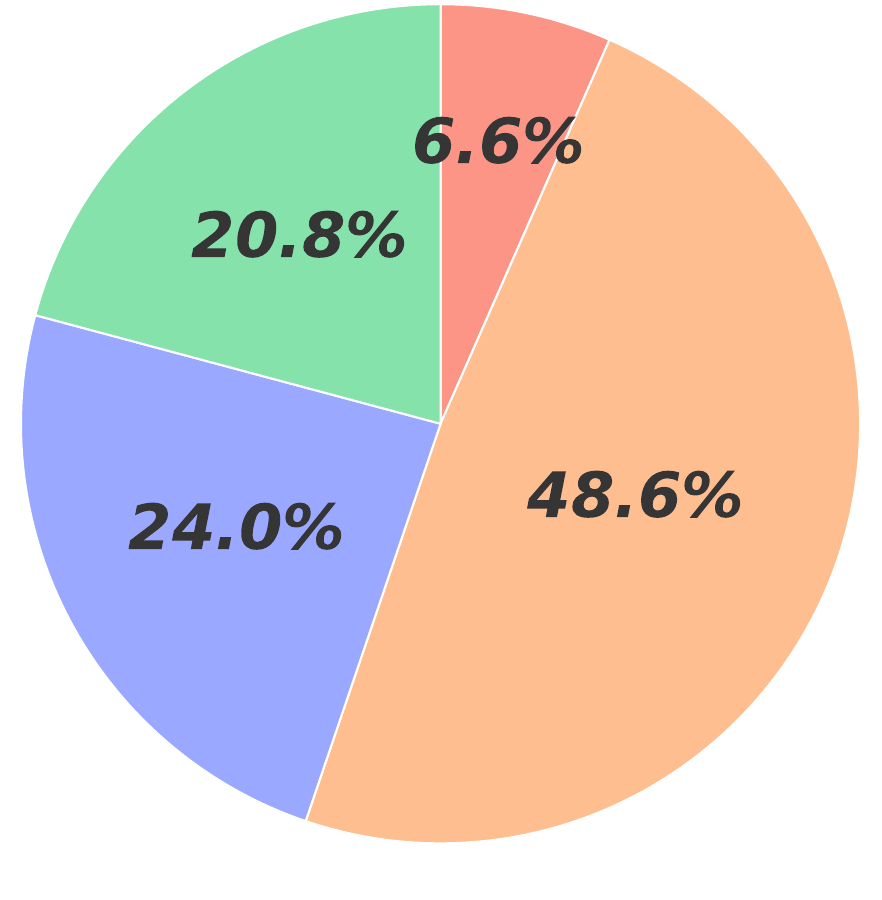}
(c) NExT-GQA (2B)
\end{minipage}
\hfill
\begin{minipage}[b]{0.22\textwidth}
\centering
\includegraphics[width=\linewidth]{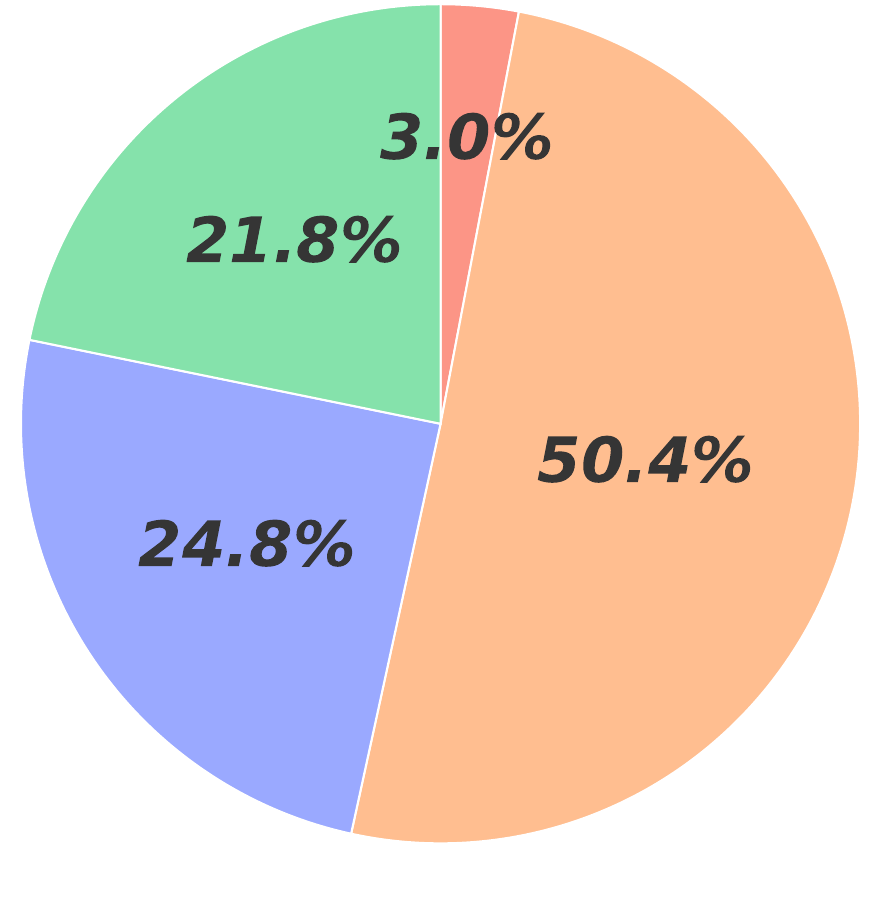}
(d) NExT-GQA (7B)
\end{minipage}
\vspace{-1mm}
\caption{Error distribution of our 2B and 7B variants on ReXTime \citep{rextime} and NExT-GQA \citep{nextgqa} datasets. The {\color{RedOrange} red}, {\color{BurntOrange} orange}, {\color{blue!70} blue}, and {\color{ForestGreen} green} portions represent planning, grounding, verification, and answering errors, respectively.}
\label{fig:error}
\vspace{-2mm}
\end{figure}

\subsection{More Detailed Analysis}

\paragraph{Timestamp Modeling Designs} The grounder plays a crucial role in our proposed Chain-of-LoRA pipeline. The model's temporal grounding quality directly impacts the final QA accuracy. To demonstrate the necessity of this design, we implement and compare the following alternative timestamp modeling techniques based on VideoMind-2B (Grounder):
\begin{enumerate}[left=0pt,topsep=0pt,itemsep=0pt,parsep=0pt]
\item \textbf{Text-only \citep{timechat}:} Directly represent timestamps in text form (\eg, ``2.3 seconds'').
\item \textbf{Special Tokens \citep{momentor}:} Define a set of timestamp tokens (\eg, \texttt{<T0>}, \texttt{<T1>}).
\item \textbf{Embedding Matching \citep{etbench}:} Predict frame features to retrieve the frame index.
\item \textbf{Time Marker \citep{timemarker}:} Explicitly insert textual timestamps among visual tokens.
\end{enumerate}
Their zero-shot video temporal grounding results are shown in Table~\ref{tab:grounder}. The results clearly demonstrate that the timestamp decoder delivers the strongest temporal grounding capability. We attribute it to two key factors: (1) It decouples continuous timestamp modeling from discrete token prediction, allowing the model to represent time with higher precision; (2) The direct regression supervision (L1 Loss) further enhances time reasoning and stabilizes training. Moreover, the timestamp decoder naturally supports predicting multiple moments with corresponding confidence scores, supporting tasks like multi-moments retrieval \citep{qvhighlights} and facilitating moment re-ranking through the verifier. These advantages jointly enhance the reliability of temporal grounding, which ensures the correct moment could be localized for further reasoning.

\begin{figure}
\centering
\hfill
\begin{minipage}[b]{0.45\textwidth}
\centering
\includegraphics[width=\linewidth]{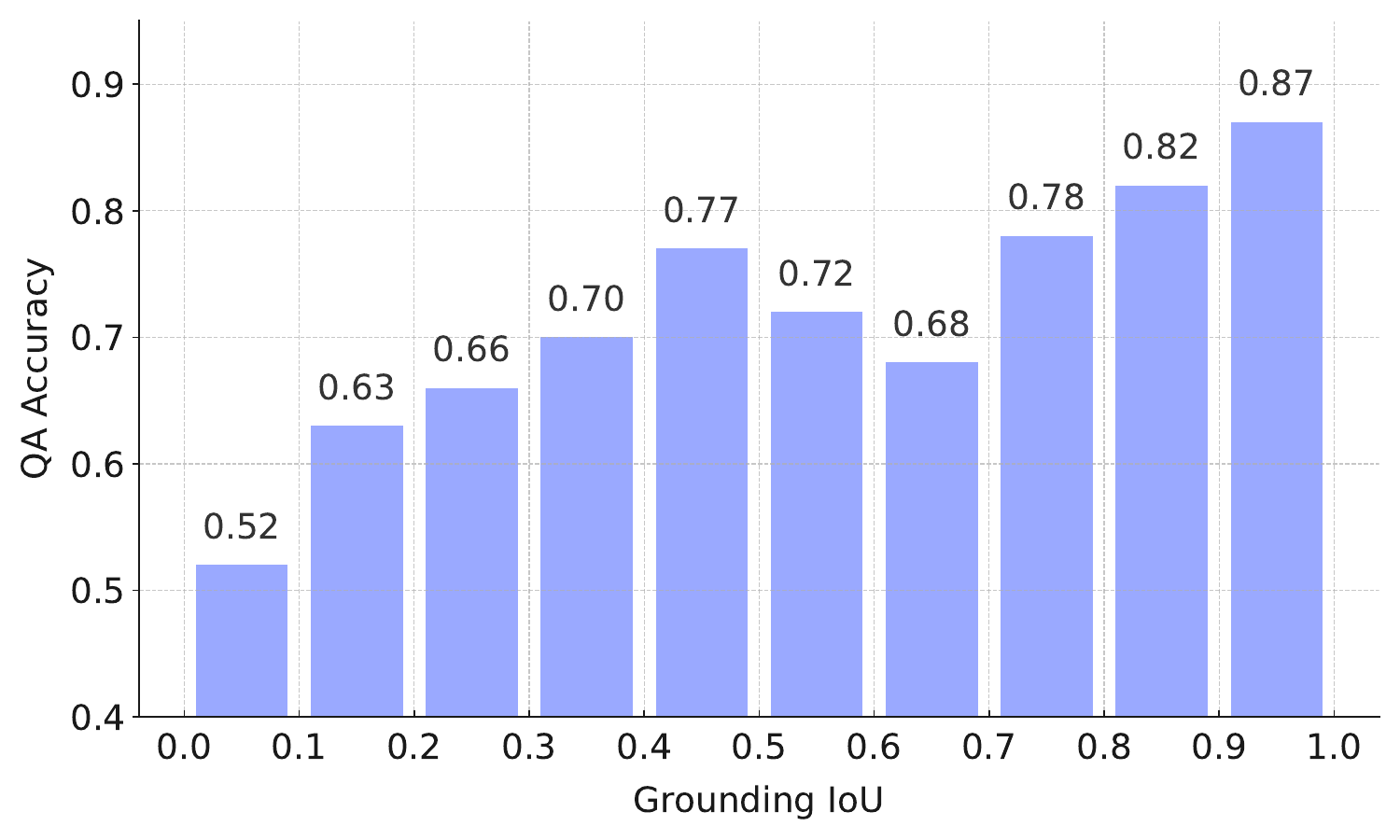}
(a) ReXTime
\end{minipage}
\hfill
\begin{minipage}[b]{0.45\textwidth}
\centering
\includegraphics[width=\linewidth]{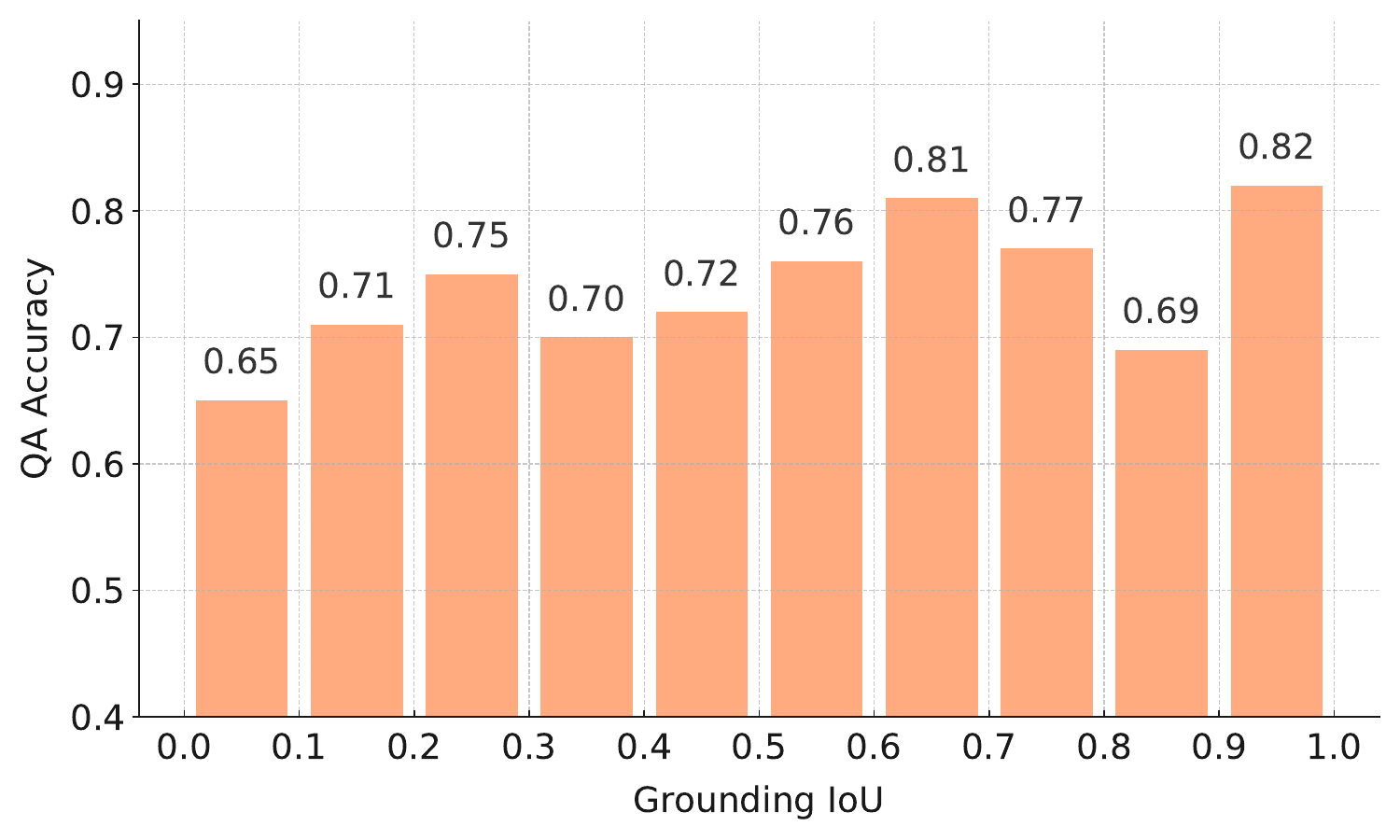}
(b) NExT-GQA
\end{minipage}
\hfill
\vspace{-1mm}
\caption{The correlation between grounding IoU and the final QA accuracy of VideoMind-2B on ReXTime \citep{rextime} and NExT-GQA \citep{nextgqa} datasets.}
\label{fig:correlation}
\vspace{-3mm}
\end{figure}

\begin{table}
\centering
\scriptsize
\begin{minipage}{0.485\textwidth}
\setlength{\tabcolsep}{6.6pt}
\caption{Effect of the temporal feature pyramid on Charades-STA \citep{charadessta}.}
\vspace{2mm}
\begin{tabularx}{\linewidth}{c|cccc}
\toprule
\multirow{2.6}{*}{\textbf{\#Pyramid Levels}} & \multicolumn{4}{c}{\textbf{Charades-STA}} \\
\cmidrule{2-5}
& R@0.3 & R@0.5 & R@0.7 & mIoU \\
\midrule
1 & 60.55 & 44.57 & 15.82 & 38.13 \\
2 & 61.51 & 46.90 & 19.36 & 40.43 \\
3 & \underline{62.62} & \underline{47.02} & \underline{20.08} & \underline{41.27} \\
\rowcolor{blue!7.5} 4 & \textbf{63.55} & \textbf{47.23} & \textbf{21.69} & \textbf{42.02} \\
\bottomrule
\end{tabularx}
\label{tab:pyramid}
\end{minipage}
\hfill
\begin{minipage}{0.485\textwidth}
\setlength{\tabcolsep}{7.65pt}
\caption{Effect of different verifier styles on Charades-STA \citep{charadessta}.}
\vspace{2mm}
\begin{tabularx}{\linewidth}{c|cccc}
\toprule
\multirow{2.6}{*}{\textbf{Verifier Type}} & \multicolumn{4}{c}{\textbf{Charades-STA}} \\
\cmidrule{2-5}
& R@0.3 & R@0.5 & R@0.7 & mIoU \\
\midrule
Direct & 60.42 & 45.28 & 19.32 & 39.84 \\
Expand & 65.10 & 48.70 & 23.15 & 43.57 \\
Textual & \underline{65.24} & \underline{49.33} & \underline{23.89} & \underline{44.01} \\
\rowcolor{blue!7.5} Special Token & \textbf{67.63} & \textbf{51.05} & \textbf{25.99} & \textbf{45.22} \\
\bottomrule
\end{tabularx}
\label{tab:verifier_design}
\end{minipage}
\vspace{-3mm}
\end{table}

\begin{table}[t!]
\centering
\scriptsize
\setlength{\tabcolsep}{3.615pt}
\caption{Effect of the verifier on Charades-STA \citep{charadessta}. \underline{IoU Raise} means the percentage of the samples whose grounding IoU is raised by the verifier.}
\vspace{2mm}
\begin{tabularx}{0.765\linewidth}{l|c|ccccc}
\toprule
\textbf{Role(s)} & \textbf{Size} & \textbf{R@0.3} & \textbf{R@0.5} & \textbf{R@0.7} & \textbf{mIoU} & \textbf{IoU Raise} \\
\midrule
Grounder & 2B & 63.2 & 46.9 & 20.5 & 41.7 & -- \\
\rowcolor{blue!7.5} Grounder + Verifier & 2B & \textbf{68.0} \textcolor{Green}{(+7.6\%)} & \textbf{51.2} \textcolor{Green}{(+9.2\%)} & \textbf{24.3} \textcolor{Green}{(+18.5\%)} & \textbf{44.8} \textcolor{Green}{(+7.4\%)} & \textbf{32.9\%} \\
\midrule
Grounder & 7B & 69.4 & 53.2 & 26.6 & 46.8 & -- \\
\rowcolor{blue!7.5} Grounder + Verifier & 7B & \textbf{73.8} \textcolor{Green}{(+6.3\%)} & \textbf{59.1} \textcolor{Green}{(+11.1\%)} & \textbf{30.1} \textcolor{Green}{(+13.2\%)} & \textbf{49.8} \textcolor{Green}{(+6.4\%)} & \textbf{31.3\%} \\
\bottomrule
\end{tabularx}
\label{tab:verifier}
\vspace{-3mm}
\end{table}

\begin{table}[t!]
\centering
\scriptsize
\begin{minipage}{0.3965\textwidth}
\setlength{\tabcolsep}{6pt}
\caption{The accuracy of planner with different input combinations.}
\vspace{2mm}
\begin{tabularx}{\linewidth}{cc|c}
\toprule
\textbf{Input Video} & \textbf{Input Question} & \textbf{Planning Acc} \\
\midrule
\cmark && 0.42 \\
& \cmark & 0.79 \\
\midrule
\rowcolor{blue!7.5} \cmark & \cmark & \textbf{0.93} \\
\bottomrule
\end{tabularx}
\label{tab:planner}
\end{minipage}
\hfill
\begin{minipage}{0.575\textwidth}
\setlength{\tabcolsep}{7.25pt}
\caption{Comparison of average inference time on CG-Bench \citep{cgbench} (avg. duration: 27 min).}
\vspace{2mm}
\begin{tabularx}{\linewidth}{l|c|c}
\toprule
\textbf{Method} & \textbf{Size} & \textbf{Inference Time (s/video)} \\
\midrule
LongVILA-R1 \citep{longvilar1} & 7B & \underline{8.75} \\
\midrule
\rowcolor{blue!7.5} \textbf{VideoMind} & 7B & 9.53 \textcolor{Red}{(+8.9\%)} \\
\rowcolor{blue!7.5} \textbf{VideoMind} (w. Auto Planning) & 7B & \textbf{8.07} \textcolor{Green}{(-7.8\%)} \\
\bottomrule
\end{tabularx}
\label{tab:inference}
\end{minipage}
\vspace{-3mm}
\end{table}

\paragraph{Effect of the Temporal Feature Pyramid} Table~\ref{tab:pyramid} studies the effectiveness of the temporal feature pyramid. Our baseline model directly makes predictions on the last-layer transformer outputs. When adding more pyramid levels, the performance of video temporal grounding consistently improves under all metrics on Charades-STA \citep{charadessta} under zero-shot setting, suggesting the effectiveness of improving the robustness of the model when facing moments with different lengths.

\paragraph{Effect of the Verifier for Zoom-in Evaluation} To quantify the verifier's corrective gain, we provide a comparison between w. and w/o the verifier on Charades-STA \citep{charadessta} in Table~\ref{tab:verifier}. The results demonstrate that the verifier consistently enhances temporal grounding performance, especially on high-quality predictions (\eg, 18.5\% higher R@0.7 on the 2B variant), highlighting its importance in the overall pipeline.

\paragraph{Design Choices of Verifier} In Table~\ref{tab:verifier_design}, we examine various design choices for the verifier. The term ``Direct'' refers to the method where the grounded moment is directly sent into the model without any expansion. ``Expand'' denotes expanding the temporal boundaries by 50\%, while ``Textual'' involves adding supplementary textual information to indicate the length of the target event. ``Special Token'' represents our final approach, utilizing special tokens to denote the grounded start and end timestamps. The comparison demonstrates that expanding the temporal boundaries effectively broadens the verifier's perceptual range, and the use of special tokens enhances the model's awareness of precise moment boundaries.

\paragraph{Reliability of the Planner} We provide an in-depth investigation into the reliability of the planner. Specifically, we randomly split the planner's training dataset into an 80\% training set and a 20\% test set, and then re-train the planner on the training set and evaluate it as a three-way classification task on the held-out test set. The metric \texttt{planning accuracy} is defined as the proportion of samples for which the predicted reasoning plan is correct. The comparison among different input combinations in Table~\ref{tab:planner} demonstrate that incorporating both video (even with low resolution) and question input substantially improves planning performance, and the resulting 93\% accuracy reflects the considerable reliability of the planner.

\begin{table}
\centering
\scriptsize
\setlength{\tabcolsep}{6.5pt}
\caption{Controlled experiments with strictly aligned hyperparameter settings. Both MLVU \citep{mlvu} and LVBench \citep{lvbench} are downsampled to 300 samples each.}
\vspace{2mm}
\begin{tabularx}{0.63\linewidth}{l|c|c|c}
\toprule
\multirow{2.6}{*}{\textbf{Method}} & \multirow{2.6}{*}{\textbf{Size}} & \textbf{MLVU (mini)} & \textbf{LVBench (mini)} \\
\cmidrule{3-4}
&& M-Avg & Overall \\
\midrule
GPT-4o \citep{gpt4o} & -- & 59.7 & 31.3 \\
Gemini-1.5-Pro \citep{gemini1.5} & -- & \underline{60.3} & \underline{36.3} \\
\midrule
\rowcolor{blue!7.5} \textbf{VideoMind} (Ours) & 2B & 59.3 & 35.7 \\
\rowcolor{blue!7.5} \textbf{VideoMind} (Ours) & 7B & \textbf{62.7} & \textbf{40.3} \\
\bottomrule
\end{tabularx}
\label{tab:control}
\vspace{-3mm}
\end{table}

\begin{table}
\centering
\scriptsize
\setlength{\tabcolsep}{5pt}
\caption{Performance comparison among the integration of our Chain-of-LoRA mechanism on different representative base models.}
\vspace{2mm}
\begin{tabularx}{0.81\linewidth}{l|c|c|c|c|c|c}
\toprule
\multirow{2.6}{*}{\textbf{Base Model}} & \multirow{2.6}{*}{\textbf{Size}} & \textbf{CG-Bench} & \textbf{ReXTime} & \textbf{Video-MME} & \textbf{MLVU} & \textbf{LVBench} \\
\cmidrule{3-7}
&& acc.@IoU & Acc@IoU & \textit{w/o sub.} & M-Avg & Overall \\
\midrule
\multirow{2}{*}{Qwen2-VL \citep{qwen2vl}} & 2B & 4.0 & 17.3 & 55.4 & 58.7 & 35.4 \\
& 7B & 4.7 & \underline{20.2} & 58.2 & \underline{64.4} & 40.8 \\
\midrule
\multirow{2}{*}{Qwen2.5-VL \citep{qwen2.5vl}} & 3B & \underline{5.0} & 15.6 & 60.9 & 62.7 & 40.5 \\
& 7B & \textbf{5.7} & 19.8 & \underline{65.9} & \textbf{66.3} & \textbf{45.2} \\
\midrule
\multirow{2}{*}{InternVL3 \citep{internvl3}} & 2B & 4.1 & 17.5 & 58.2 & 61.4 & 38.1 \\
& 8B & 4.5 & \textbf{20.8} & \textbf{66.5} & 63.8 & \underline{42.3} \\
\bottomrule
\end{tabularx}
\label{tab:lmm}
\vspace{-3mm}
\end{table}

\begin{table}[t!]
\centering
\scriptsize
\setlength{\tabcolsep}{6.5pt}
\caption{Performance of the simulated multi-role pipelines on closed-source models. Both MLVU \citep{mlvu} and LVBench \citep{lvbench} are downsampled to 300 samples each.}
\vspace{2mm}
\begin{tabularx}{0.69\linewidth}{l|c|c|c}
\toprule
\multirow{2.6}{*}{\textbf{Method}} & \multirow{2.6}{*}{\textbf{\makecell{Multi-Role\\ Pipeline}}} & \textbf{MLVU (mini)} & \textbf{LVBench (mini)} \\
\cmidrule{3-4}
&& M-Avg & Overall \\
\midrule
\multirow{2}{*}{GPT-4o \citep{gpt4o}} & \xmark & 59.7 & 31.3 \\
& \cmark & \textbf{62.3} \textcolor{Green}{(+4.4\%)} & \textbf{32.7} \textcolor{Green}{(+4.5\%)} \\
\midrule
\multirow{2}{*}{GPT-5 \citep{gpt5}} & \xmark & 61.7 & 34.3 \\
& \cmark & \textbf{63.3} \textcolor{Green}{(+2.6\%)} & \textbf{36.3} \textcolor{Green}{(+5.8\%)} \\
\midrule
\multirow{2}{*}{Gemini-2.5-Pro \citep{gemini2.5}} & \xmark & 73.3 & 65.7 \\
& \cmark & \textbf{76.3} \textcolor{Green}{(+4.1\%)} & \textbf{68.7} \textcolor{Green}{(+4.6\%)} \\
\bottomrule
\end{tabularx}
\label{tab:api}
\vspace{-3mm}
\end{table}

\paragraph{Inference-Time Efficiency} In Table~\ref{tab:inference}, we study the inference-time efficiency of our method on CG-Bench \citep{cgbench}. All experiments are conducted on a single NVIDIA RTX 6000 Ada GPU. Compared with the text-based reasoning baseline LongVILA-R1 \citep{longvilar1}, our full pipeline is approximately 8.9\% slower. However, this gap can be easily bridged by activating the planner's auto-planning capability. When the planner is allowed to choose the reasoning path, some easy questions are routed directly to the answerer, which substantially reduces the average inference time from 9.53s to 8.07s per video, resulting 7.8\% faster inference speed than the baseline.

\paragraph{Overall Robustness and Error Accumulation} We acknowledge that the proposed sequential reasoning pipeline has the potential risk of error propagation and accumulation. To quantify this effect, we conduct a systematic analysis of error propagation on two representative datasets: ReXTime \citep{rextime} (more temporal-related) and NExT-GQA \citep{nextgqa} (more reasoning-related). For both datasets, each error case is categorized into one of the following types: (1) \textbf{Planning Error:} The question can only be correctly answered by switching to another reasoning plan (\eg, from ``all roles'' to ``answerer only''); (2) \textbf{Grounding Error:} All the top-5 predicted moments are incorrect (\ie, having temporal IoU $<$ 0.5); (3) \textbf{Verification Error:} The moment selected after verification is incorrect; (4) \textbf{Answering Error:} The predicted answer is incorrect.

We present the case distributions in Table~\ref{tab:case} and error distributions in Figure~\ref{fig:error}. Several conclusions can be drawn from the results: (1) The planner is highly reliable, accounting for less than 5\% of the error cases on both datasets; (2) Grounding errors account for roughly half of all failures. This is aligned with our hypothesis that accurate temporal grounding plays a crucial role in the multi-role reasoning pipeline; (3) Verification and answering contribute comparably smaller portions of the failures, accounting for only about 20\% error cases each.

\paragraph{Correlation between Grounding IoU and QA Accuracy} We study the correlation between temporal grounding performance and QA accuracy in Figure~\ref{fig:correlation}. Specifically, we group the samples in ReXTime \citep{rextime} and NExT-GQA \citep{nextgqa} datasets into different IoU buckets, and plot the average QA accuracy within each bucket. On ReXTime, which is more temporal-related, the results show a clear positive correlation between grounding IoU and final QA accuracy. On the more reasoning-related NExT-GQA, such correlation is less significant.

\paragraph{Controlled Experiments on Closed-source APIs} The results of closed-source models in Table~\ref{tab:cgbench} and Table~\ref{tab:general} are reported from the corresponding benchmark papers, without strictly aligned settings. Therefore, we provide a controlled experiment to validate the advantages of our method. Specifically, we select two challenging long video understanding benchmarks, \ie, MLVU \citep{mlvu} and LVBench \citep{lvbench}, and randomly sample 300 QA pairs from each, forming MLVU (mini) and LVBench (mini). We align the key hyperparameters as follows:
\begin{enumerate}[topsep=0pt,itemsep=0pt,parsep=0pt]
\item \textbf{Frame Rate:} 1 FPS
\item \textbf{Max Frame Count:} 150
\item \textbf{Frame Resolution:} max 448 $\times$ 448 pixels with natural aspect ratio
\item \textbf{Model Hyperparameters:} \texttt{temperature\;=\;0}, \texttt{top\_p\;=\;0}, \texttt{top\_k\;=\;0}
\end{enumerate}
The comparisons are presented in Table~\ref{tab:control}, clearly showing that our VideoMind-7B outperforms both GPT-4o \citep{gpt4o} and Gemini-1.5-Pro \citep{gemini1.5} on both datasets.

\paragraph{Integration with More Open-source LMMs} In Table~\ref{tab:lmm}, we study whether the proposed Chain-of-LoRA pipeline provides a consistent benefit across different base models. The results show that when integrated with stronger base models like Qwen2.5-VL \citep{qwen2.5vl} and InternVL3 \citep{internvl3}, the performance of our Chain-of-LoRA pipeline could be further enhanced on multiple long video benchmarks, highlighting our method's generalizability.

\paragraph{Integration with Closed-source LMMs} We are also interested in whether the proposed multi-role pipeline could be simulated via a series of prompts on closed-source models. To investigate this, we evaluate the effectiveness of the multi-role reasoning prompt when applied to three models, \ie, GPT-4o \citep{gpt4o}, GPT-5 \citep{gpt5}, and Gemini-2.5-Pro \citep{gemini2.5}, on the previously constructed MLVU (mini) and LVBench (mini). The results in Table~\ref{tab:api} show that our pipeline consistently boosts the performance of different closed-source models, highlighting an interesting finding: the multi-role reasoning pipeline itself can systematically enhance long video reasoning, even without role-specific model designs or training.

\section{Miscellaneous}

\subsection{Prompt Templates}\label{sec:prompt}

We present the prompts used in this work, including the input prompts for each role of VideoMind and the prompt for GPT-4o mini \citep{gpt4o} for data annotation.

\vspace{1mm}
\noindent\textbf{Prompt for the Planner:}
\begin{tcolorbox}[colback=plannercolor,boxrule=0pt,colframe=white,rounded corners,left=2mm,right=2mm,top=2mm,bottom=2mm,arc=3mm]
\footnotesize
You are acting as the planner now. Given a question about the video, your task is to analyze the question and identify the best way to answer this question. You have access to the following tools: \\\\
Grounder: Accepts a text query and localizes the relevant video segment according to the query. \\
Verifier: A tool supporting grounder by verifying the reliability of its outputs. \\
Answerer: Answer a given question directly based on the whole video or a cropped video segment. \\\\
Your response must be a list in JSON format. A valid plan for reasoning could be ``grounder, verifier, answer'', ``grounder, verifier'', or ``answerer'', depending on the given question. Please see an example of the format below. \\\\
\text{[\{``type'': ``grounder'', ``value'': ``text query''\}, \{``type'': ``verifier''\}, \{``type'': ``answerer''\}]} \\\\
Note that only the grounder can accept an argument called ``value'', which is the text query used for grounding. Now I give you the question: ``\textbf{\{question\}}''. Please think carefully and respond with your plan in JSON directly.
\end{tcolorbox}

\vspace{1mm}
\noindent\textbf{Prompt for the Grounder:}
\begin{tcolorbox}[colback=groundercolor,boxrule=0pt,colframe=white,rounded corners,left=2mm,right=2mm,top=2mm,bottom=2mm,arc=3mm]
\footnotesize
You are acting as the grounder now. Given a video and a text query, your goal is to temporally localize the video moment described by the query. If the query is directly describing a moment, simply localize it according to its content. Otherwise, if the moment is described as ``before/after a pivotal event'', you need to determine the actual event it refers to. The localized moment should only cover the target event. Now I give you the query: ``\textbf{\{query\}}''. Please think carefully and provide your response.
\end{tcolorbox}

\vspace{1mm}
\noindent\textbf{Prompt for the Verifier:}
\begin{tcolorbox}[colback=verifiercolor,boxrule=0pt,colframe=white,rounded corners,left=2mm,right=2mm,top=2mm,bottom=2mm,arc=3mm]
\footnotesize
You are acting as the verifier now. You will be presented a text query describing a moment that potentialy happens in the given video. Your task is to identify whether the video segment between \texttt{<SEG-START>} and \texttt{<SEG-END>} perfectly covers the moment. If the described moment can be seen in the video, please focus on verifying whether the moment starts at \texttt{<SEG-START>} and ends at \texttt{<SEG-END>}. Respond with ``Yes'' if you think the moment boundaries are correct, otherwise ``No''. If the described moment cannot be seen in the video, respond with ``No'' directly. Now I give you the query: ``\textbf{\{query\}}''. Please think carefully and respond with ``Yes'' or ``No'' directly.
\end{tcolorbox}

\vspace{1mm}
\noindent\textbf{Prompt for the Answerer:} When subtitles are considered, we only present the first 100 lines.
\begin{tcolorbox}[colback=answerercolor,boxrule=0pt,colframe=white,rounded corners,left=2mm,right=2mm,top=2mm,bottom=2mm,arc=3mm]
\footnotesize
You are given a video with \textbf{\{duration\}} seconds long. \\
Subtitles: \textbf{\{subtitles\}} \\
\textbf{\{question\}} \\
Options: \\
(A) \textbf{\{option 1\}} \\
(B) \textbf{\{option 2\}} \\
(C) \textbf{\{option 3\}} \\
(D) \textbf{\{option 4\}} \\
Please only give the best option.
\end{tcolorbox}

\vspace{1mm}
\noindent\textbf{Prompt for Query Rephrasing Data Generation:}
\begin{tcolorbox}[colback=promptcolor,boxrule=0pt,colframe=white,rounded corners,left=2mm,right=2mm,top=2mm,bottom=2mm,arc=3mm]
\footnotesize
You are an expert in rewriting questions into queries. I will give you a question that requires to be answered based on a specific moment in a video. Your task is to analyze the question and rewrite it into a declarative sentence, which could be used as a text query to search for the relevant video moment. The query should be concise, describing the key event or key scene that the question asks for. \\\\
Here are some examples: \\\\
Question: How does the male cyclist react when he sees the steep path? \\
Query: The male cyclist sees the steep path. \\\\
Question: What did the girl do at the end of the video? \\
Query: The end of the video. \\\\
Question: What did the lady do as she was cycling off? \\
Query: The lady is cycling off. \\\\
Question: What is the person with red shirt doing on the yacht? \\
Query: The person with red shirt stays on the yacht. \\\\
\mbox{Now I give you the question: ``\textbf{\{question\}}''. Please think carefully and respond with the query directly.}
\end{tcolorbox}

\section{The Use of LLMs Statement}

Large Language Models (LLMs) were used in this study to aid in polishing the manuscript. Specifically, we used LLMs to assist in refining the language and detecting potential grammatical errors. This is to improve readability and ensure clarity of the paper. We confirm that LLMs were not involved in research ideation, method exploration, and experiment designs. All research ideas, methods, and analysis were produced by the authors. We take full responsibility for the content in this paper, including the text generated or polished by the LLMs.

\end{document}